\documentclass[runningheads]{llncs}
\usepackage{graphicx}
\usepackage{amsmath,amssymb} 
\usepackage{color}
\usepackage[width=122mm,left=12mm,paperwidth=146mm,height=193mm,top=12mm,paperheight=217mm]{geometry}
\usepackage{float}
\usepackage[percent]{overpic}
\usepackage{subfig}
\usepackage{hhline}
\usepackage[font=footnotesize]{caption}

\newcommand{\tab}{\hspace*{2em}}

\newcommand{\norm}[1]{\left\| {#1} \right\|}

\newcommand{\set}[1]{\left\{ {#1} \right\}}

\newcommand{\R}{\ensuremath \mathbb{R}}

\newcommand{\Y}{\mathcal{Y}}

\renewcommand{\epsilon}{\varepsilon}

\def\utilde#1{\mathord{\vtop{\ialign{##\crcr
$\hfil\displaystyle{#1}\hfil$\crcr\noalign{\kern1.5pt\nointerlineskip}
$\hfil\tilde{}\hfil$\crcr\noalign{\kern1.5pt}}}}}

\newcommand{\raquel}[1]{{\color{red}{#1}}}

\newcommand{\todo}[1]{{\color{green}{#1}}}

\newcommand{\hide}[1]{}

\begin{document}
\pagestyle{headings}
\mainmatter
\def\ECCV16SubNumber{1167}  

\title{Soccer Field Localization from a Single Image} 



\author{Namdar Homayounfar$^{\dag,\star}$, Sanja Fidler$^\dag$, Raquel Urtasun$^\dag$\\
\tab\\
{\small $^\dag$Department of Computer Science, University of Toronto}\\
{\small $^\star$Department of Statistical Sciences, University of Toronto}\\
{\tt \small \{namdar,fidler,urtasun\}@cs.toronto.edu}}
\institute{}

\authorrunning{N. Homayounfar, S. Fidler, R. Urtasun}

\maketitle

\begin{abstract}
In this work, we propose a novel way of efficiently localizing a soccer field from a single broadcast image of the game. Related work in this area relies on manually annotating a few key frames
   and extending the localization to similar images, or installing
   fixed specialized cameras in the stadium from which the layout of
   the field can be obtained. In contrast, we formulate this problem
   as a branch and bound inference in a Markov random field where an
   energy function is defined in terms of field cues such as grass,
   lines and circles. Moreover, our approach is fully automatic and
   depends only on single images from the broadcast video of the
   game. We demonstrate the effectiveness of our method by applying
   it to various games and obtain promising results. Finally, we
   posit that our approach can be applied easily to other sports such
   as hockey and basketball.
\end{abstract}

\section{Introduction}

According to  recent studies\footnote{\url{http://www.researchmoz.us/sports-analytics-market-shares-strategy-and-//forecasts-worldwide-2015-to-2021-report.html}}, the sport analytics market was worth 125 million dollars in 2014.  Current predictions expect it to reach 4.7 billion dollars by 2021. 
Sport analytics is used to increase the team's competitive edge by gaining insight into the different aspects of its playing style and the performance of each of its players. For example, sports analytics was a major component of Germany's successful World Cup 2014 campaign \hide{(http://blogs.wsj.com/cio/2014/07/10/germanys-12th-man-at-the-world-cup-big-data/)}. 
Another important application of sports analytics is to improve scouting by identifying talented prospects in junior leagues and assessing their competitive capabilities and potential fit in a future team's roster. 
Sports analytics are also beneficial in fantasy leagues, giving fantasy players access to statistics that can enhance their game play. 
Even more impressive is the global sports betting market, which is worth up to trillion dolloars according to Statista\footnote{http://www.statista.com/topics/1740/sports-betting/}. One can imagine the value of an algorithm that can predict who will win a particular match. 




Core to most analytics is the ability to automatically extract valuable information from video. Being able to identify team formations and strategies as well as assessing the performance of individual players is reliant upon understanding where the actions are taking place in 3D space.  

Most approaches to player detection \cite{Okuma2004,Tong2011,Okuma2013,Lu2013}, game event recognition \cite{Gao2011}, and team tactical analysis \cite{Niu2012,Franks2015,Liu2006}
perform field localization by either semi-manual methods \cite{Kim2000,Yamada2002,Farin2003,Watanabe2004,Fei2007,Gupta2011a,Okuma2004a,Dubrofsky2008,Hess2007} or by obtaining the game data from fixed and calibrated camera systems installed around the venue.

\begin{figure}[t]
\vspace{-0.5cm}
\centering
\includegraphics[scale=0.30]{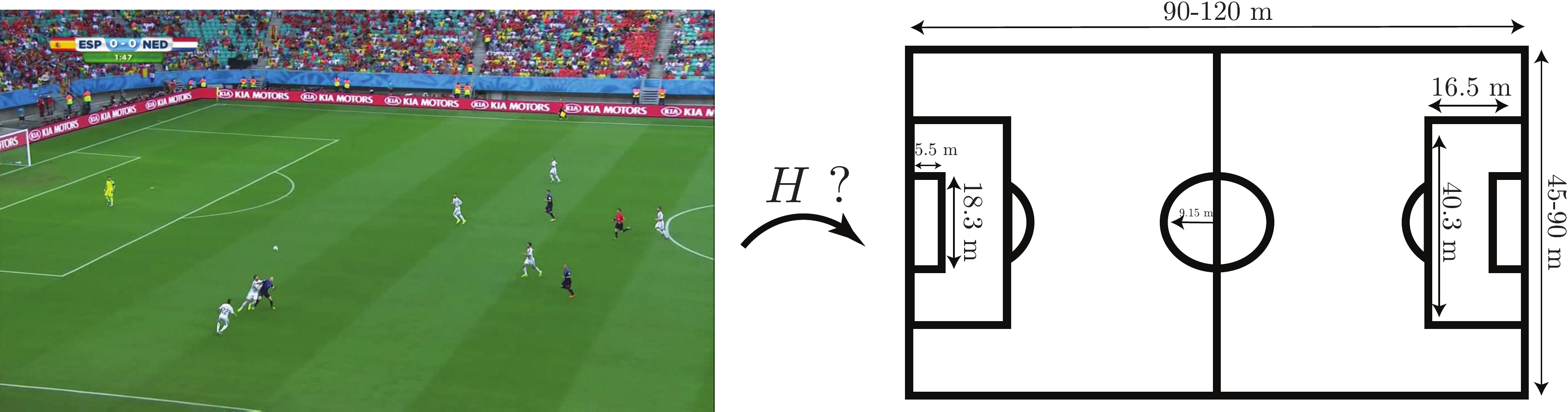}
\vspace{-0.3cm}
\caption{We seek to find the mapping $H$ (a homography) that takes the image to the geometric model of the field.}
\vspace{-0.3cm}
\label{fig:motivation}
\end{figure}

In this paper, we tackle the challenging task of field localization as applied to a single broadcast image.  We propose
a method that requires no manual initialization and is applicable to any video of the game recorded with a single camera. The input to our system is a single image and the 3D model of the field, and the output is the mapping that takes the image to the model as illustrated  in Fig. \ref{fig:motivation}.  In particular, we frame the field localization problem as inference in a Markov Random Field. We parametrize the field in terms of four rays, cast from two automatically detected horizontal vanishing points. The rays correspond to the outer lines of the field and thus define the field's precise localization. 
Our MRF energy uses several potentials that exploit semantic segmentation of the image in terms of ``grass'', as well as agreement between the lines found in the image and those defined by the known model of the field. All of our potentials can be efficiently computed. We perform inference with branch-and-bound, achieving on average 0.7 seconds running time per frame. The weights in our MRF are learned using structure SVM \cite{Tsochantaridis2005}.



We focus our efforts in the game of soccer as it is more challenging than other sports, such as  hockey or basketball. A hockey rink or a basketball court are much smaller compared to a soccer field and are  in a closed venue. In contrast, a soccer field is usually in an open stadium exposed to different weather and lightning conditions which might create difficulties in identifying the important markings of the field. Furthermore, the texture and pattern of the grass in a soccer field differs from one stadium to another in comparison to say a hockey rink which is always white. We note however that our method is sports agnostic and is easily extendable as long as the sport venue has known dimensions and primitive markings such as lines and circles. 

To evaluate our method, we collected a dataset of 259 images from 12 games in the World Cup 2014. We report the Intersection over Union (IOU) scores of our method against the ground truth, and show very promising results. 
In the following, we start with a discussion of related literature, and then describe our method. Experimental section provides an exhaustive evaluation of our method, and we finish with a conclusion and a discussion of future work.

\vspace{-3mm}
\section{Related Work}

A variety of approaches  have been developed  in industry and academia to tackle the field localization problem.  
In the industrial setting, companies such as Pixelot and Prozone have proposed a hardware approach to field localization by  developing advanced calibrated camera systems that are installed in a sporting venue. 
This requires expensive equipment, which is only possible at the highest performance level. 
Alternatively,  companies such as Stathleates rely entirely on human workers for establishing the homography between the field and the model for every frame of the game.  

In the academic setting, the common approach to field registration is to first initialize the system by either searching over a large parameter space (e.g. camera parameters) or by manually establishing a homography for various representative keyframes of the game and then propagating this homography throughout the  consecutive frames. 
In order to avoid accumulated errors, the system needs to be reinitialized by manual intervention. Many methods have been developed which exploit geometric primitives such as lines and/or circles to estimate the camera parameters\cite{Kim2000,Yamada2002,Farin2003,Watanabe2004,Fei2007}. These approaches rely on hough transforms or RANSAC and require manually specified color and texture heuristics. 

An approach to limit the search space of the camera parameters is to find the two principal vanishing points corresponding to the field lines~\cite{Hayet2004a,Hayet2007} and only look at the lines and intersection points that are in accordance with these vanishing points and which satisfy certain cross ratios. The efficacy of the method was demonstrated only on goal areas where there are lots of visible lines. However, this approach faces problems for views of the centre of the field, where there are usually fewer lines and thus one cannot   estimate the  vanishing point reliably. 



In~\cite{Suat2007}, the authors proposed an approach that matches images of the game to 3D models of the stadium for initial camera parameter estimation \cite{Suat2007}. However, these 3D models only exist in well known stadiums, limiting the applicability of the proposed approach.  

Recent approaches, applied to Hockey, Soccer and American Football \cite{Gupta2011a,Okuma2004a,Dubrofsky2008,Hess2007} require a manually specified homography for a representative set of  keyframe images per recording. 
In contrast, in this paper we propose a method that only relies on images taken from a single camera. Also no temporal information or manual initialization is required. Our approach could be used, for example in conjunction with \cite{Gupta2011a,Okuma2004a} to automatically produce smooth high quality field estimates from video.

\begin{figure}[t]
\vspace{-0.5cm}
     \centering
\subfloat[]{
	\includegraphics[width=0.5\linewidth]{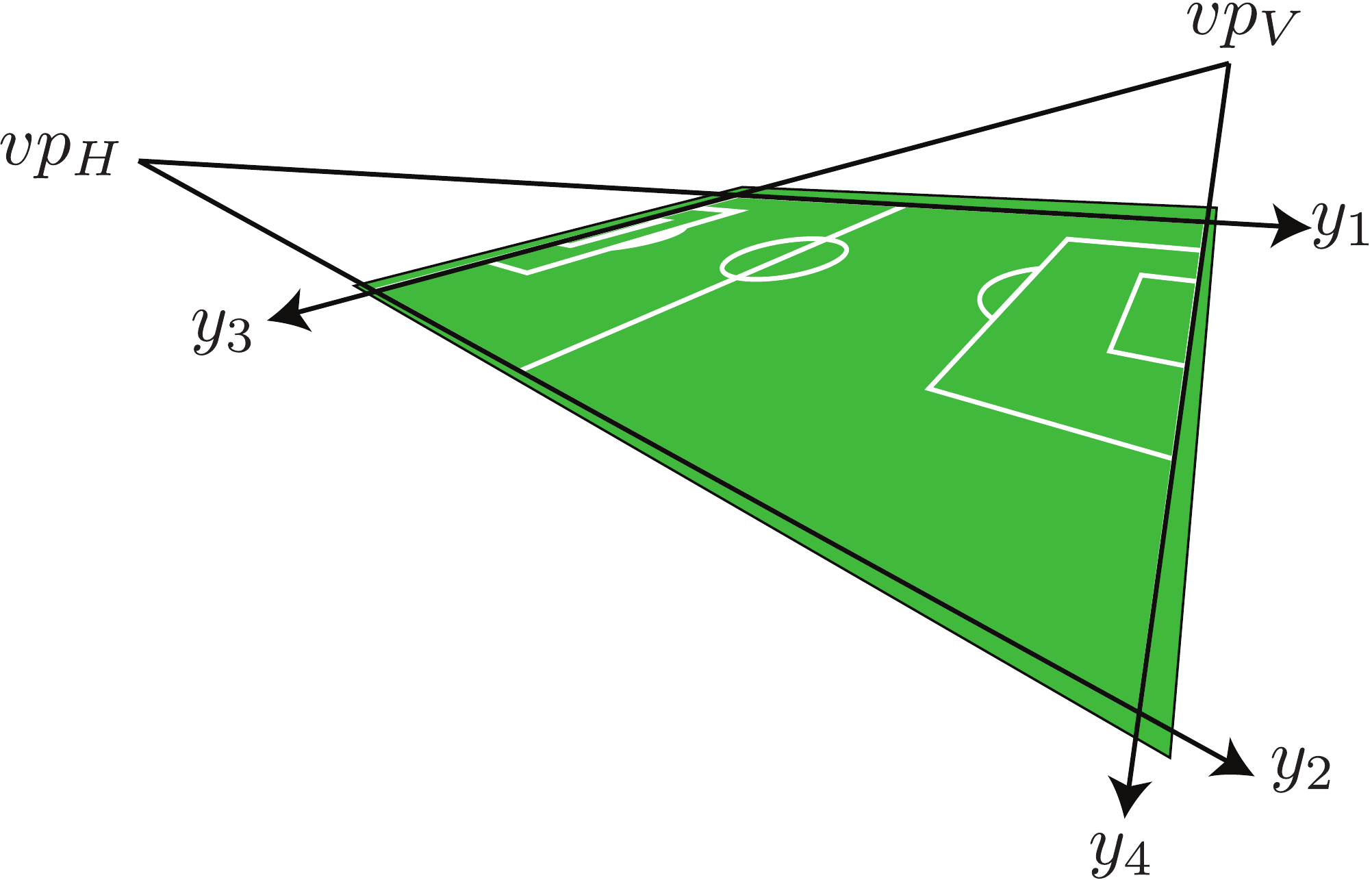}
	 }
\subfloat[]{
	\includegraphics[width=0.5\linewidth]{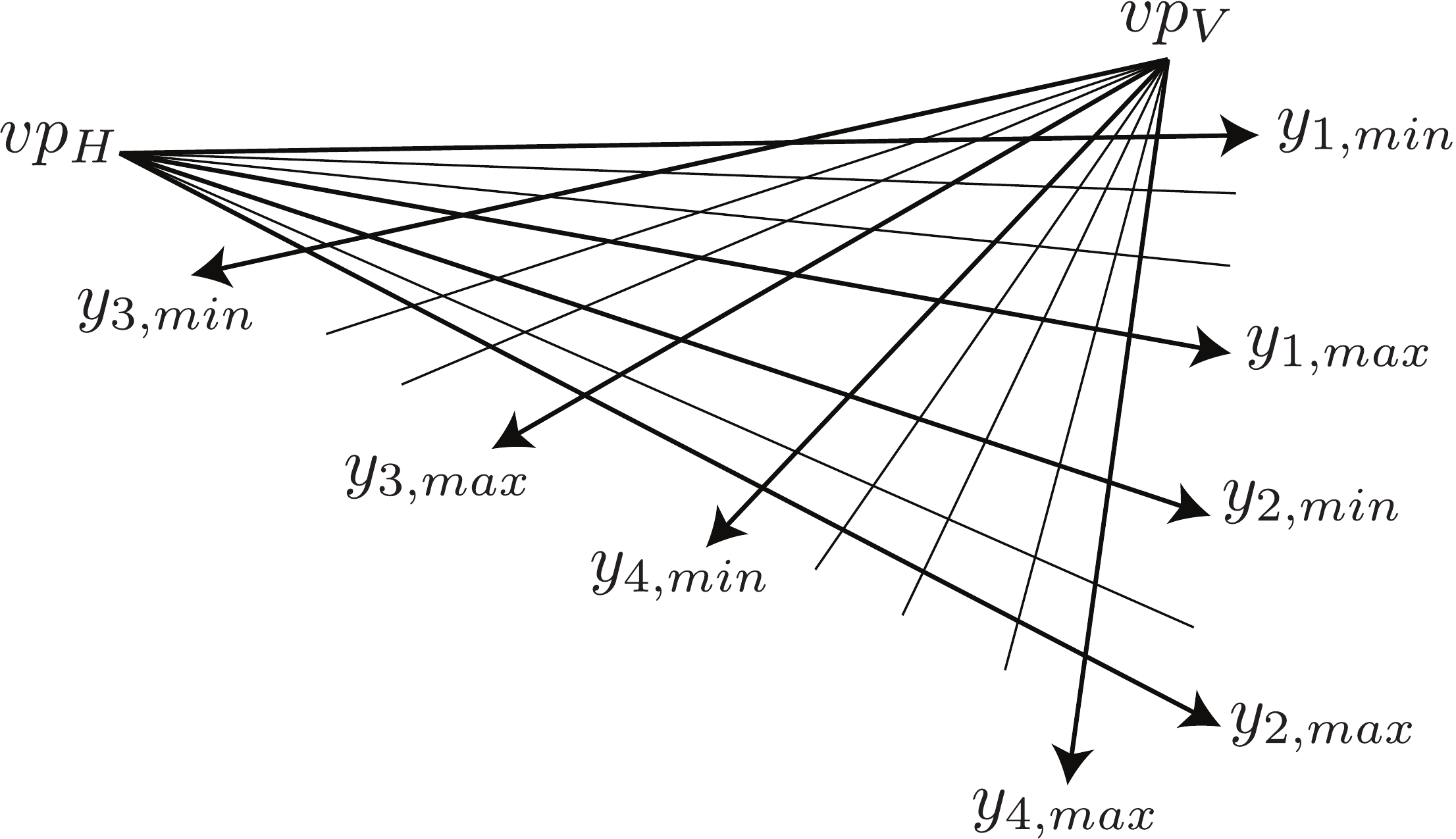}
  }
  \vspace{-0.3cm}
     \caption{(a) Field parametrization in terms of 4 rays $y_i$. (b) The grid 
     }
     \label{fig:field_image}
     \vspace{-1mm}
\end{figure}

\section{3D Soccer Field Registration}

The goal of this paper is to automatically compute the transformation between a broadcast image of a soccer field, and the 3D geometric  model of the field. 
In  this section, we first show how to parameterize the problem by making use of the vanishing points, reducing the effective number of degrees of freedom to be estimated. We then formulate the problem as energy minimization in a Markov random field that encourages agreement between the model and the image in terms of grass segmentation as well as the location of  the primitives (i.e., lines and ellipses) that define the soccer field.  Furthermore, we show that inference can be solved exactly very efficiently via branch and bound.

\vspace{-2mm}
\subsection{Field Model and Parameterization}

Assuming that the ground is planar, a soccer field can be represented by a 2D rectangle embedded in a 3D space. 
The rectangle can  be defined by two long line segments referred to as touchlines and two shorter line segments, each behind a goal post, referred to as goallines. 
Each soccer field has also a set of vertical and horizontal lines defining the goal areas, the penalty boxes, and the midfield. \hide{\raquel{ what else?}} Additionally,  a full circle and two semicircles are also highlighted which define distances that opposing players should maintain from the ball at kickoff \hide{\raquel{ define this as well}}. We refer the reader to Fig. \ref{fig:motivation} for an illustration of the geometric field model.

The transformation between the field in the broadcast image and our 3D model can be parameterized with a homography $H$, which is a $3\times 3$ invertible matrix defining a bijection that maps lines to lines between 2D projective spaces  \cite{Hartley2004}. The matrix $H$ has 8 degrees of freedom and encapsulates the transformation of the broadcast image to the soccer field model.
A common way to estimate this homography is by detecting points and lines in the image and associating them with points and lines in the soccer field model. Given these correspondences, the homography can be estimated in closed form using the Direct Linear Transform (DLT) algorithm \cite{Hartley2004}. While a closed form solution is very attractive, the problem lies on the fact that the association of lines/points between the image and the soccer model is not known a priori. Thus, in order to solve for the homography, one needs to evaluate all possible assignments. 
 As a consequence DLT-like algorithms are  typically used in the scenario where a nearby solution is already known (from a keyframe or previous frame), and search is done over a small set of possible associations. 
 
 \begin{figure}[t]
\vspace{-0.4cm}
     \centering
\subfloat[]{
	\includegraphics[width=0.5\linewidth]{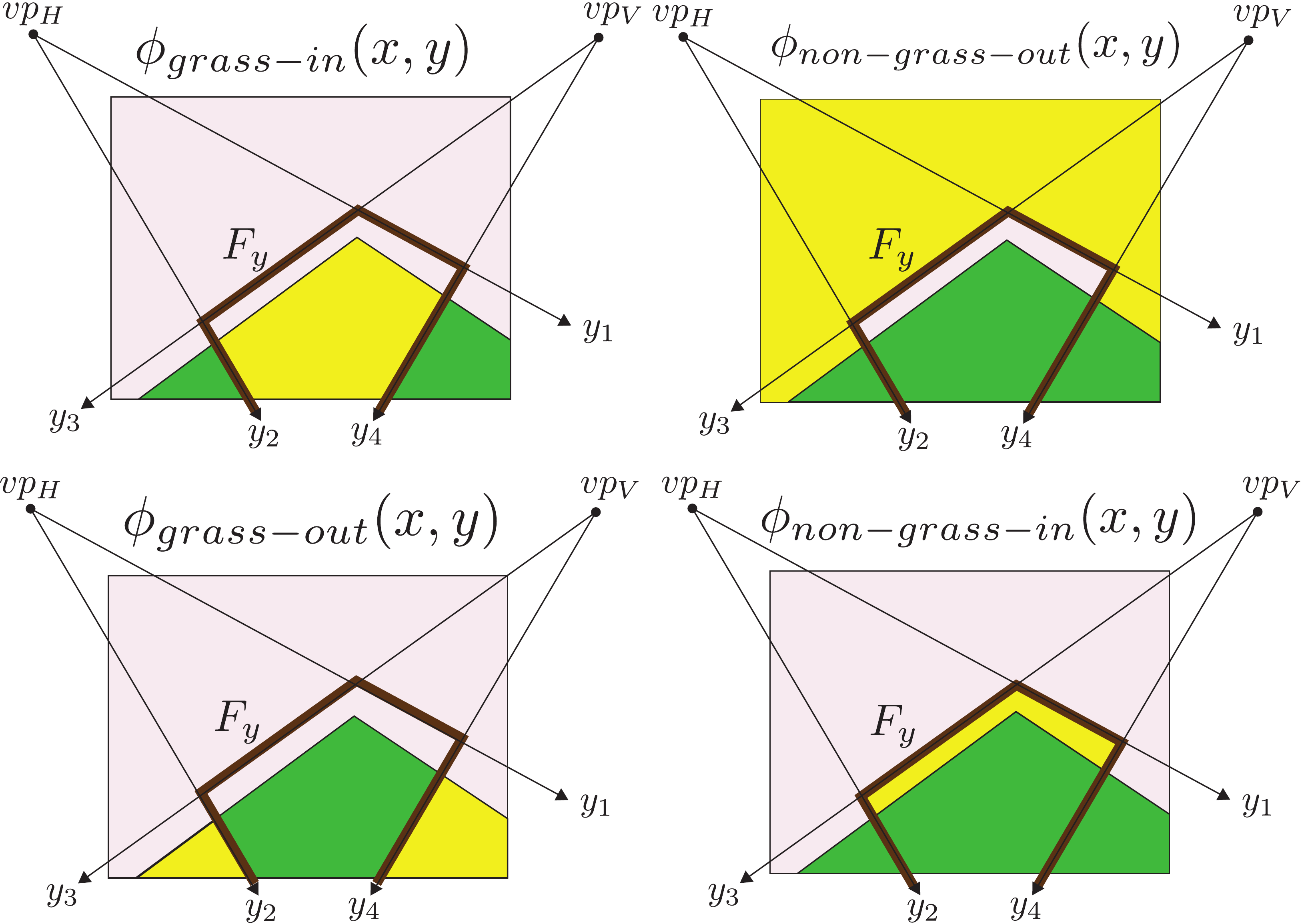}
	 }
\subfloat[]{
	\includegraphics[width=0.5\linewidth]{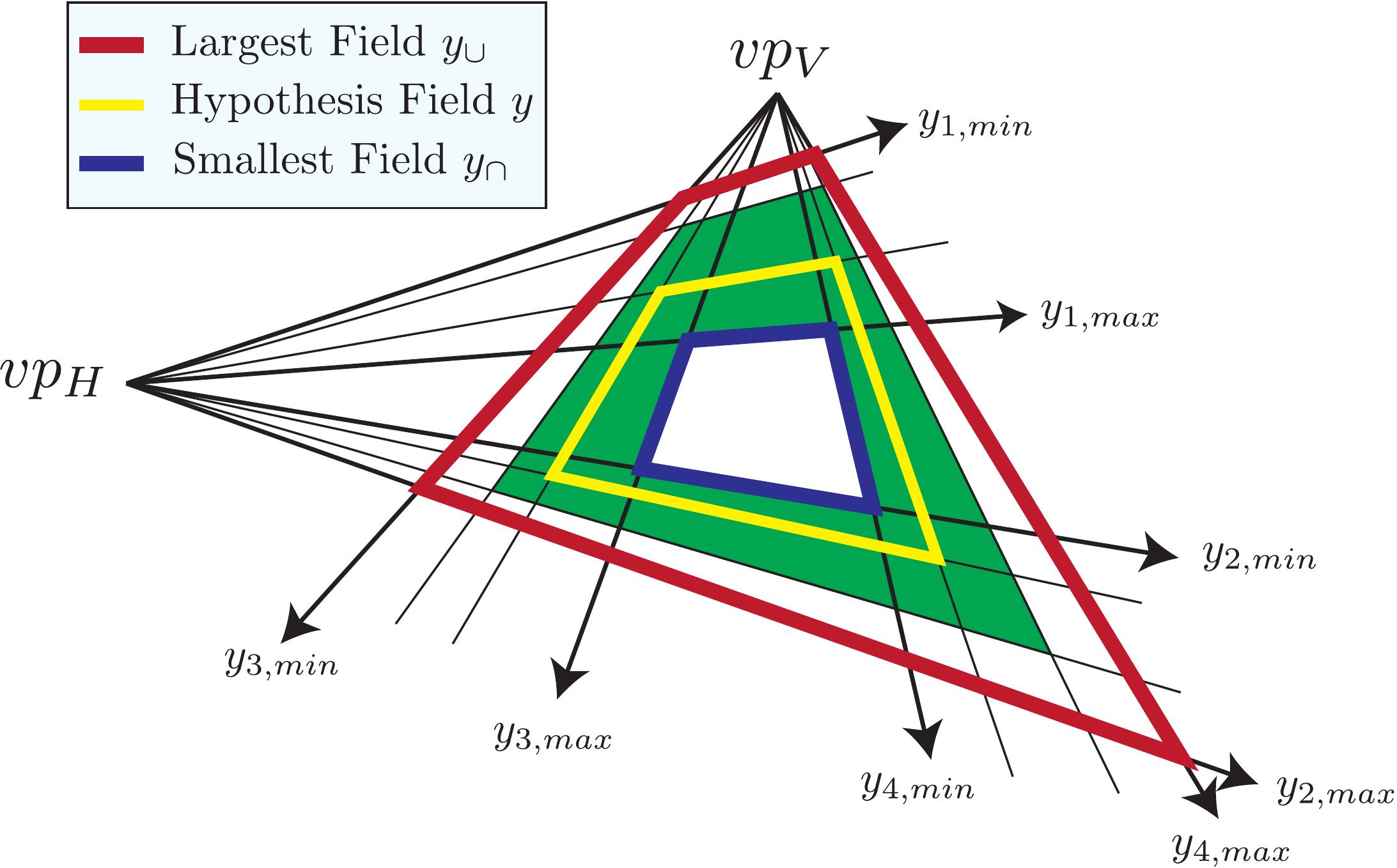}
  }
  \vspace{-4mm}
     \caption{(a) In each plot, the green area correspond to grass and the grey area to non-grass pixels. The field $F_y$ is the region inside the highlighted lines. The yellow region is the percentage of counted grass/non-grass pixels. (b) The red line is the largest possible field and the blue line is the smallest field.} 
     \label{fig:grass_image}
\end{figure}
 

 In this paper, we follow a very different approach, which jointly solves  for the association and the estimation of the homography. Towards this goal, we first reduce the effective number of degrees of freedom of the homography. 
In an image of the field,  parallel lines intersect at two orthogonal vanishing points.  If we can estimate the vanishing points reliably we can reduce the number of degree of freedom from 8 to 4. We defer the discussion about how we estimate the vanishing points to section \ref{sec:VP}. 



For  convenience of presentation, we refer to the lines parallel to the touchlines as horizontal lines, and the lines parallel to the goallines as vertical lines. Let  $x$ be an image of the field. Denote by $vp_{V}$ and  $vp_{H}$ the (orthogonal) vertical and horizontal vanishing points respectively. 
 Since a football stadium conforms to a Manhattan world, there also exists a third vanishing point which is orthogonal to both $vp_{V}$ and $vp_{H}$. We omit this third vanishing point from our model since there are  usually not many lines enabling us to compute it reliably. 

We define a hypothesis field by four rays  emanating from the  vanishing points. The rays $y_1$ and $y_2$ originate from $vp_{H}$ and correspond to the touchlines. Similarly, the rays $y_3$ and $y_4$ originate from $vp_{V}$ and  correspond to the goallines. As depicted in Fig. \ref{fig:field_image}, a hypothesis field is constructed by the intersection of the four rays. 
Let the tuple $y = (y_1, \dots, y_4) \in \Y$ 
be the parametrization of the field, where  we have discretized the set of possible candidate rays. Each ray $y_i$  falls in an  interval $[y_{i,min}^{init}, y_{i,max}^{init}]$ and $\Y=\prod_{i=1}^4\set{[y_{i,min}^{init}, y_{i,max}^{init}]}$ is the product space of these four integer intervals. 
Thus $\Y$ corresponds to a grid.  



\vspace{-2mm}
\subsection{Field Estimation as Energy Minimization}

In this section, we parameterize the problem as the one of inference in a Markov random field. In particular, given an image $x$ of the field, we obtain the best prediction $\hat{y}$  by solving the following inference task:
\begin{equation}
	\hat{y} = \arg\max_{y \in \Y}\, w^T \phi (x,y) \label{eq:inference}
\end{equation}
with $\phi(x,y)$ a feature vector encoding various potential functions and $w$ the set of corresponding weights which we learn using structured SVMs \cite{Tsochantaridis2005}. 
In particular, our energy defines different potentials encoding the fact that the field should contain mostly grass, and high scoring configurations prefer  the projection of the field primitives (i.e., lines and circles) to  be aligned with the detected primitives in the image (i.e. detected line segments and conic edges). 
In the following we discuss the potentials in more detail.


\vspace{-3mm}
\subsubsection{Grass Potential:} This potential encodes the fact that a soccer field is made of grass. 
We perform semantic segmentation of the broadcast image into grass vs. non-grass. Towards this goal,  we exploit the prediction from a CNN trained using DeepLab \cite{chen14semantic} for our binary segmentation task. 
Given a hypothesis field $y$, let $F_y$ denote the field restricted to the image $x$.
We would like to maximize the number of grass pixels in $F_{y}$. Hence, we define a potential function, denoted by $\phi_{grass-in}(x,y)$, that counts the percentage of total grass pixels that fall inside the hypothesis field $F_y$. However, note that for any  hypothesis $y'$ with   $F_{y} \subset F_{y'}$,  $F_{y'}$ would have at least as many  grass pixels as $F_{y}$. This introduces a bias towards hypotheses that correspond to zoom-in cameras. 
We thus define three additional  potentials such that we try to minimize the number of grass pixels outside the field $F_y$ and the number of non-grass pixels inside  $F_y$, 
while maximizing the number of non-grass pixels outside $F_{y}$. We denote these potentials as $\phi_{grass-out}(x,y)$, $\phi_{non-grass-out}(x,y)$ and $\phi_{non-grass-in}(x,y)$ respectively. 
We refer the reader to Fig. \ref{fig:grass_image} for an illustration. 



\vspace{-3mm}
\subsubsection{Lines Features:} 
\begin{figure}[t]
\vspace{-5mm}
     \centering
\subfloat[]{
	\includegraphics[width=0.24\linewidth]{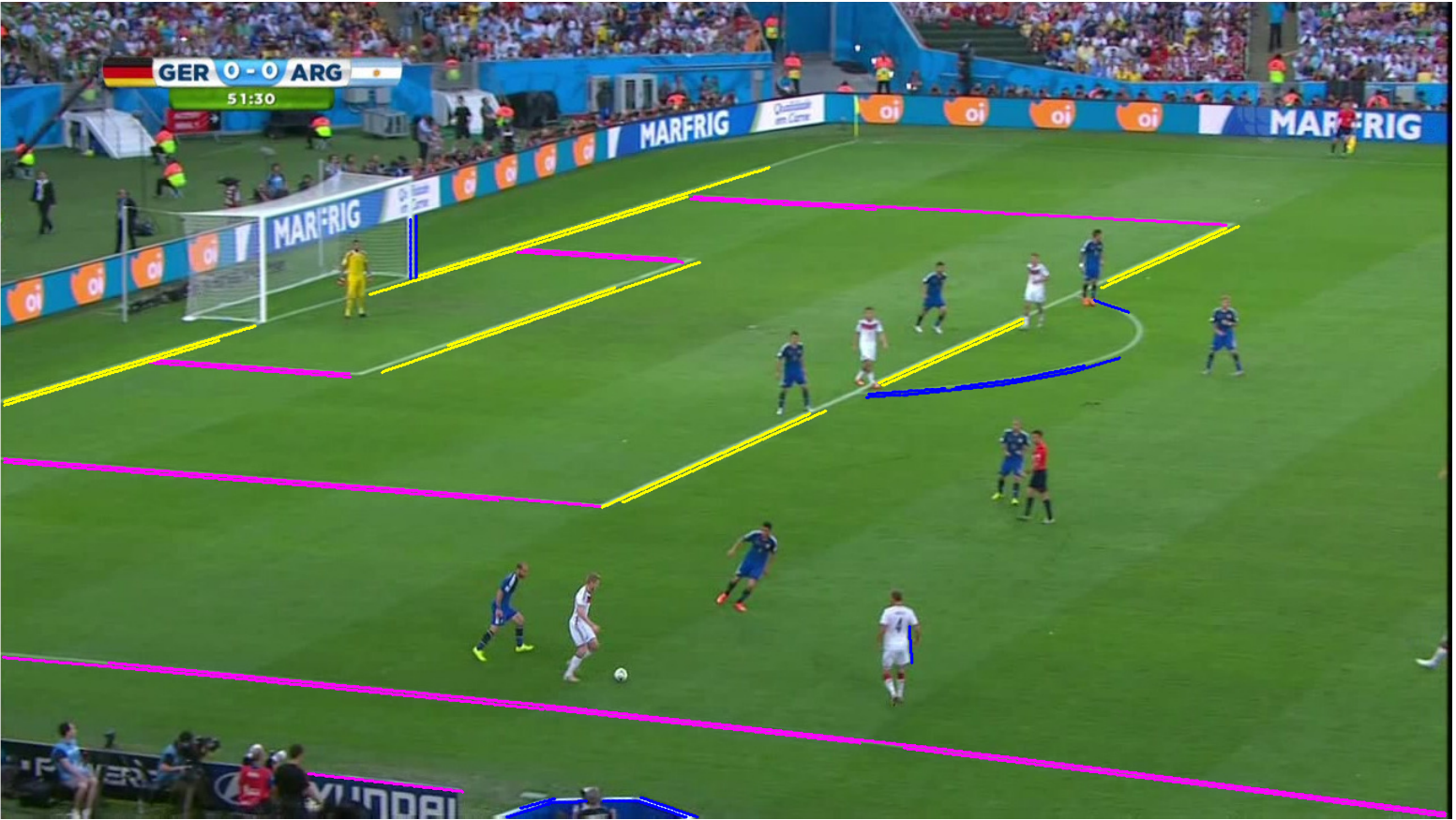}
	 }
\subfloat[]{
	\includegraphics[width=0.24\linewidth]{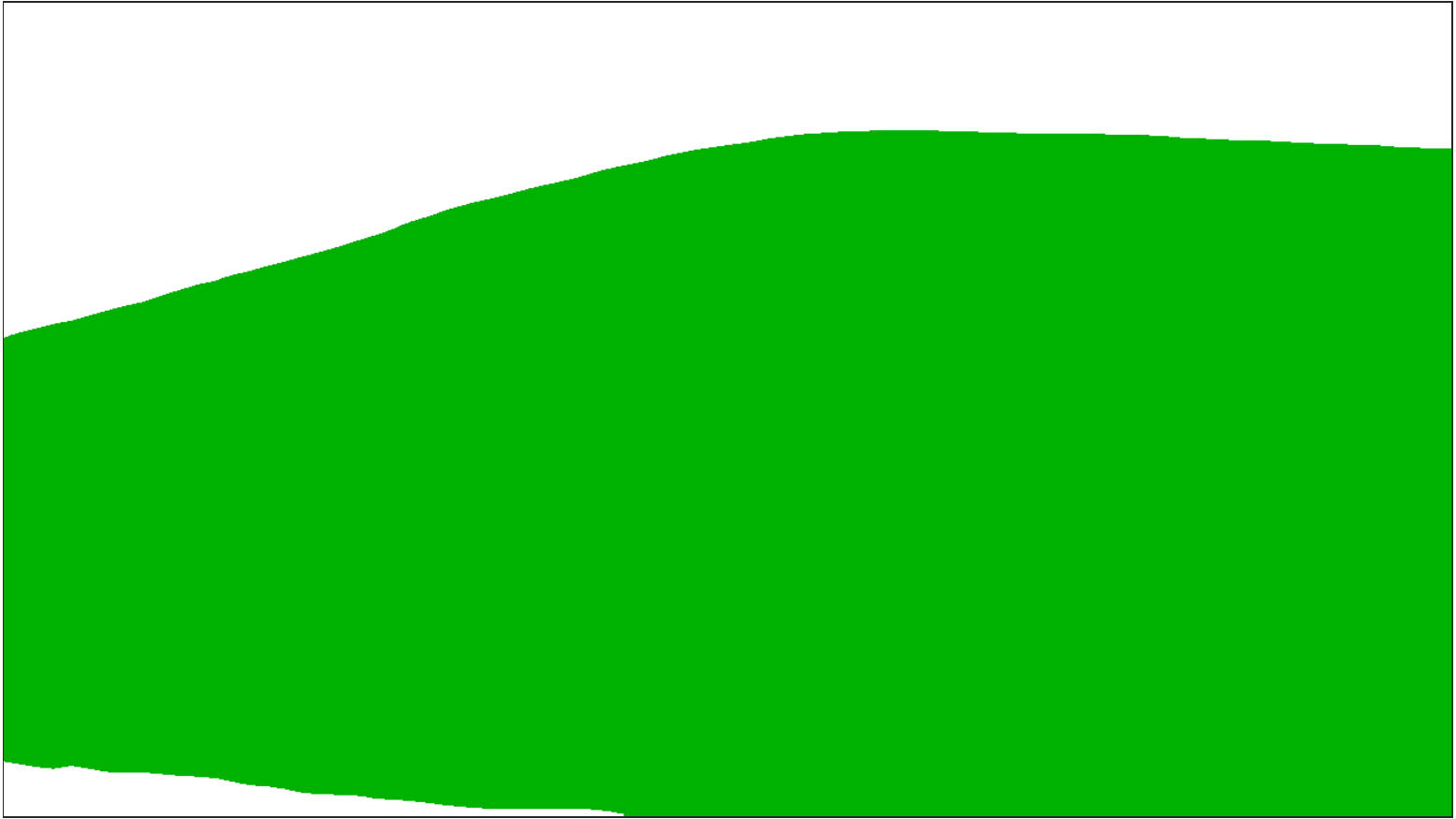}
  }
  \subfloat[]{
	\includegraphics[width=0.24\linewidth]{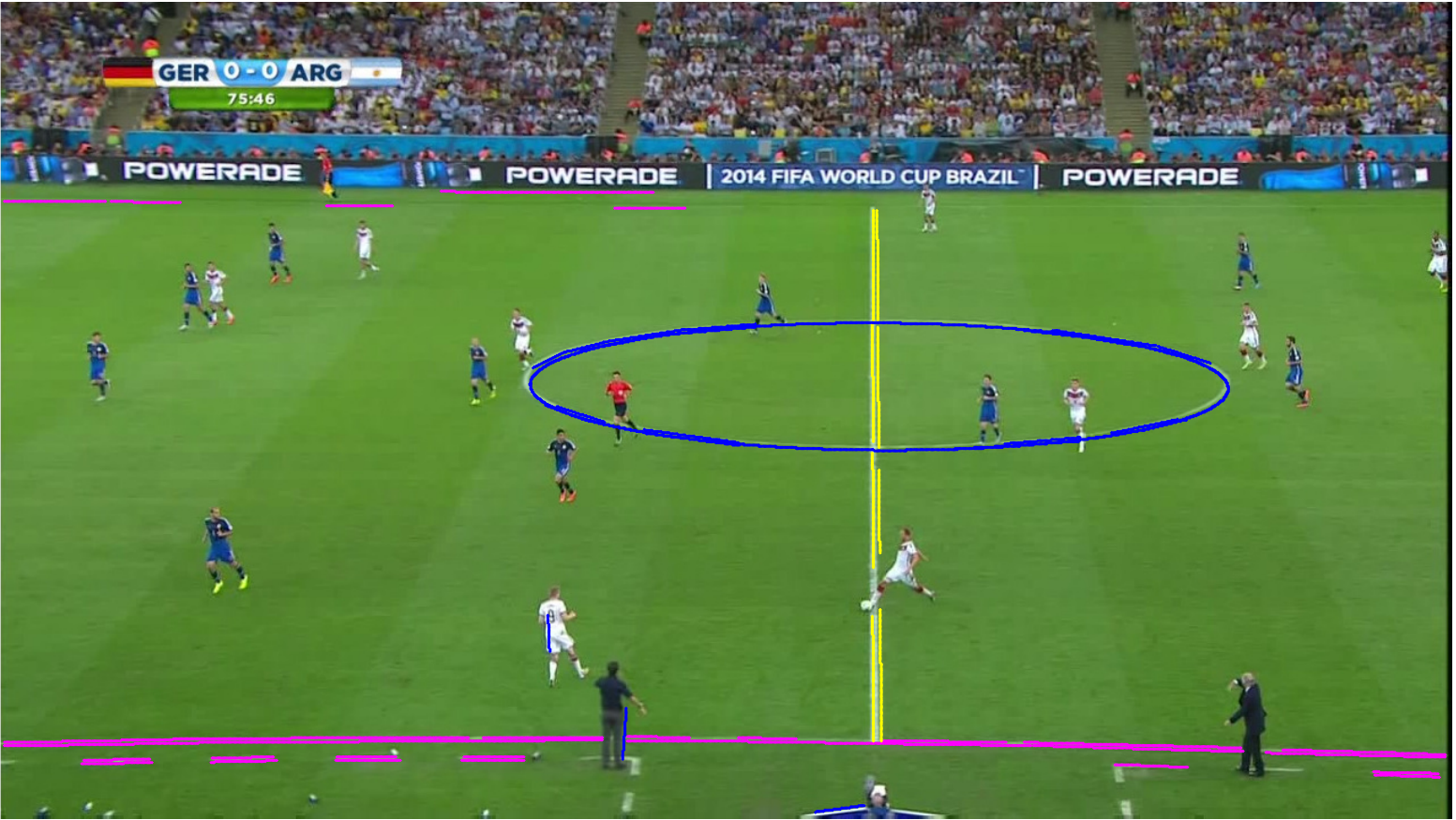}
	 }
\subfloat[]{
	\includegraphics[width=0.24\linewidth]{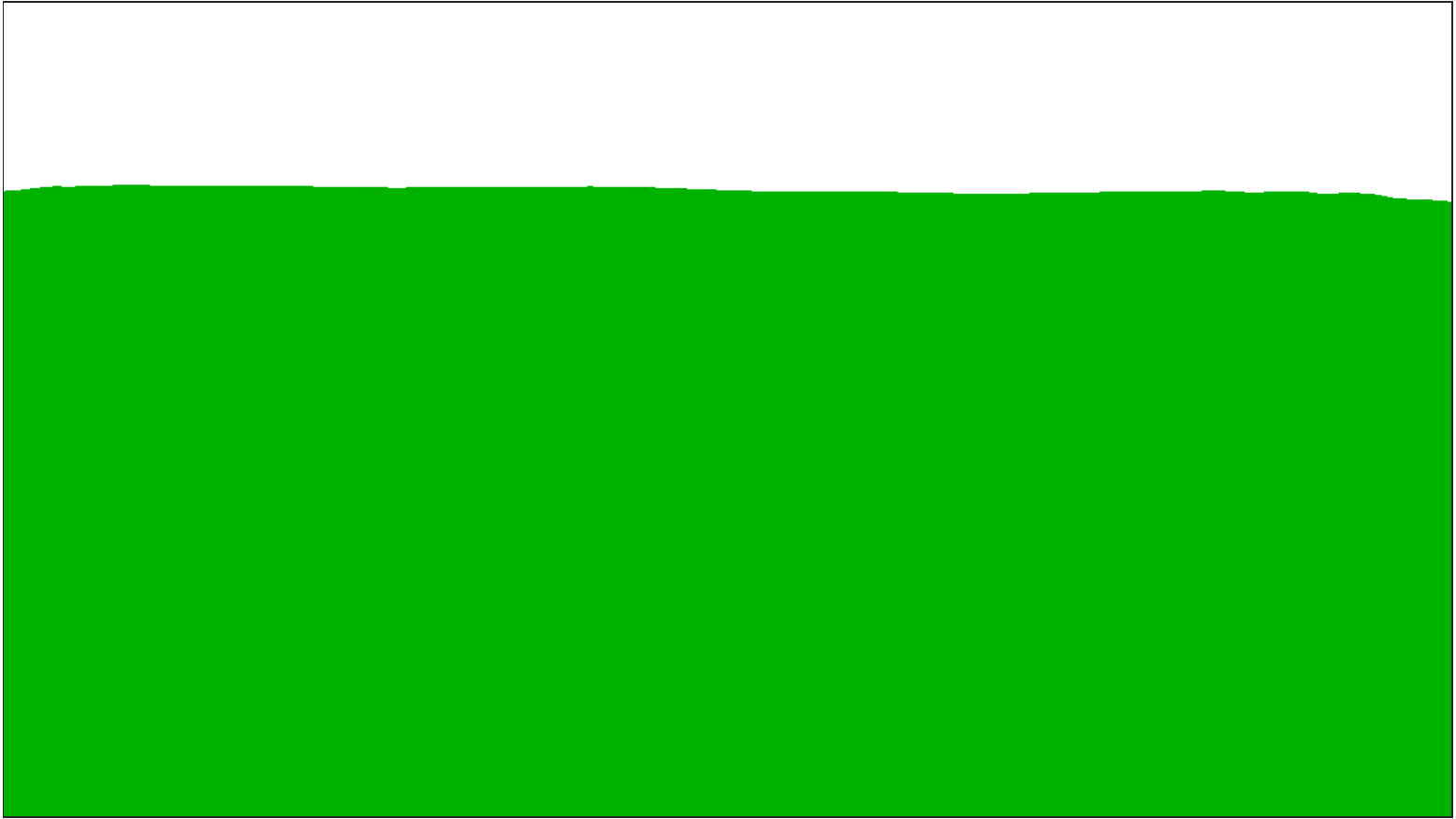}
  }
  \vspace{-3mm}
     \caption{(a),(c) Two images of the game. The detected yellow and magenta line segments correspond to $vp_V$ and $vp_H$ respectively. The blue line segments do not correspond to any vanishing point. (b),(d) The grass segmentation results for the images in (a)/(c)} 
     \label{fig:vp_grass}
     \vspace{-3mm}
\end{figure}

\begin{figure}[t]
\vspace{-0.6cm}
     \centering
\subfloat[]{
	\includegraphics[width=0.66\linewidth]{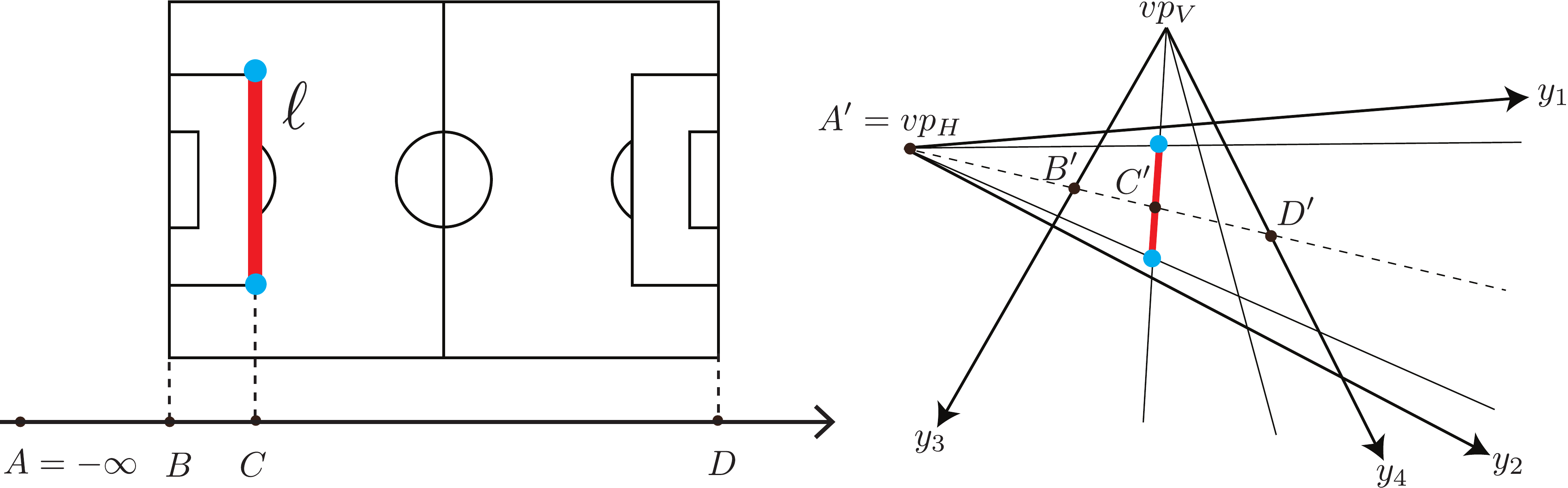}
	 }
\subfloat[]{
	\includegraphics[width=0.33\linewidth]{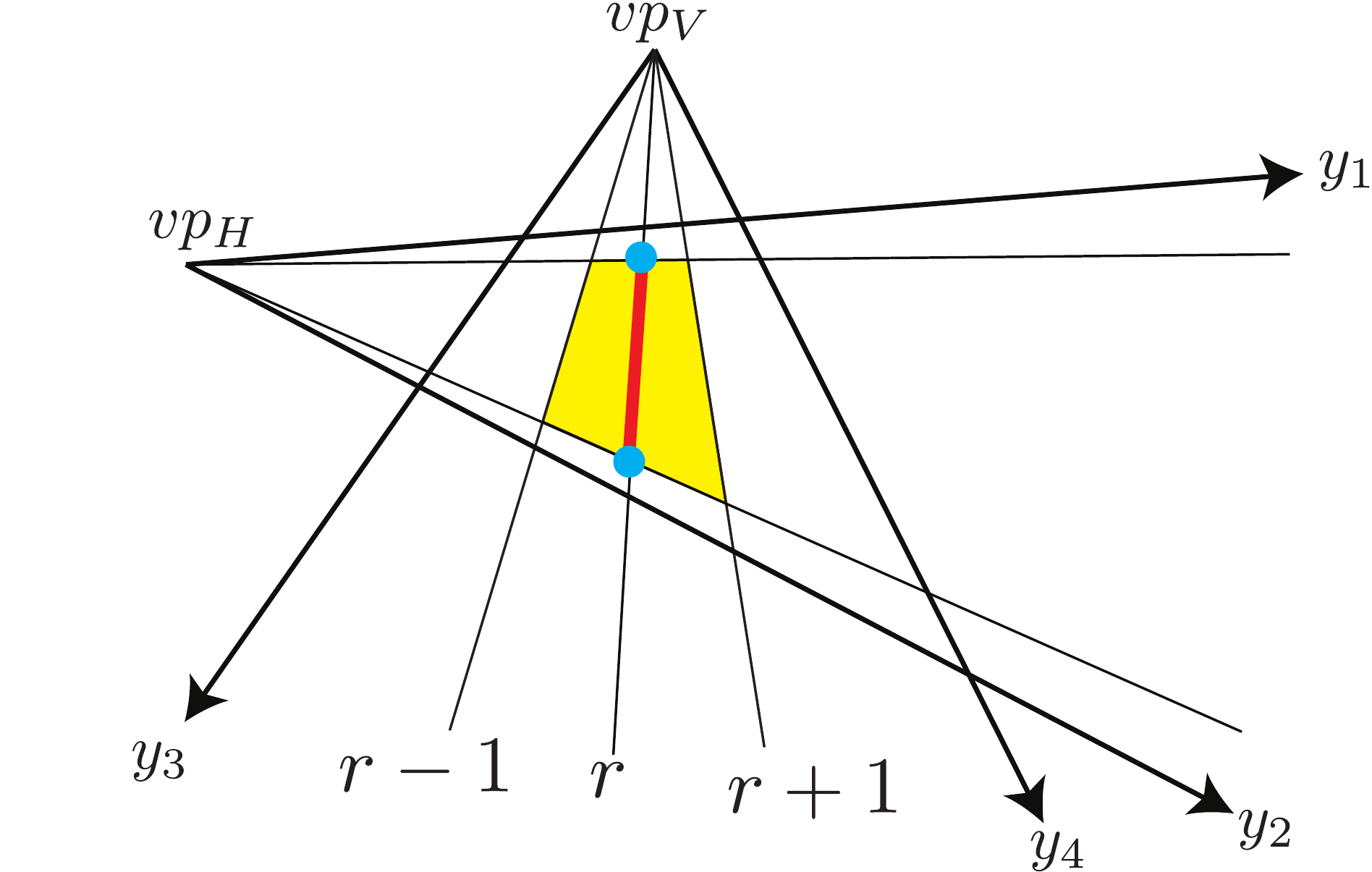}
  }
  \vspace{-0.3cm}
     \caption{(a) For line $\ell$ (red line) in the model, the cross ratio $CR = BD/BC$ must equal the cross ratio of the projection of $\ell$ on the grid given by $CR' = (A'C' \cdot B'D')/(BC'\cdot A'D')$. The projection of the endpoints of $\ell$ are computed similarly. (b) For vertical line $\ell$, the potential $\phi_{\ell}(x,y)$ counts the percentage of $vp_V$ line pixels in the yellow region for which the vertical sides are one ray away from the ray on which $\ell$ falls upon.} 
     \label{fig:cross-ratios}
\end{figure}

The observable lines corresponding to the white marking of the soccer field provide strong clues on the location of the touchlines and goallines. This is because their positions and lengths must always  adhere to the FIFA specifications. In a soccer field there are 7 vertical and 10 horizontal line segments as depicted in Fig.~\ref{fig:motivation}.
Using the line detector of~\cite{Rafael2012}, we find all the line segments in the image and also the vanishing points as described in section~\ref{sec:VP}. A byproduct of our vanishing point estimation  procedure is that each detected line segment is assigned to   $vp_H$, $vp_V$  or  none (e.g. line segments that fall on the ellipse edges) as demonstrated in Fig. \ref{fig:vp_grass}. 
We then define a scoring function $\phi_{\ell_i}(x,y)$ for each line $\ell_i$, $i=1,\dots,17$ that is large when the image evidence agrees with the predicted line position obtained by reprojecting the model using the hypothesis $y$. 
The exact reprojection can be easily obtained by using the invariance property of cross ratios~\cite{Hartley2004}, Fig. \ref{fig:cross-ratios}(a).
Giving  the exact position of a line $\ell_i$ on the grid $\Y$, the score $\phi_{\ell_i}(x,y)$ counts the percentage of line segment pixels that are aligned with the same vanishing point, Fig. \ref{fig:cross-ratios}(b). 
We refer the reader to the suppl. material for more in  details.




\vspace{-3mm}
\subsubsection{Circle Potentials:}
\label{sec:circle}
\begin{figure}[t]
\vspace{-0.2cm}
\centering
\includegraphics[scale=0.30]{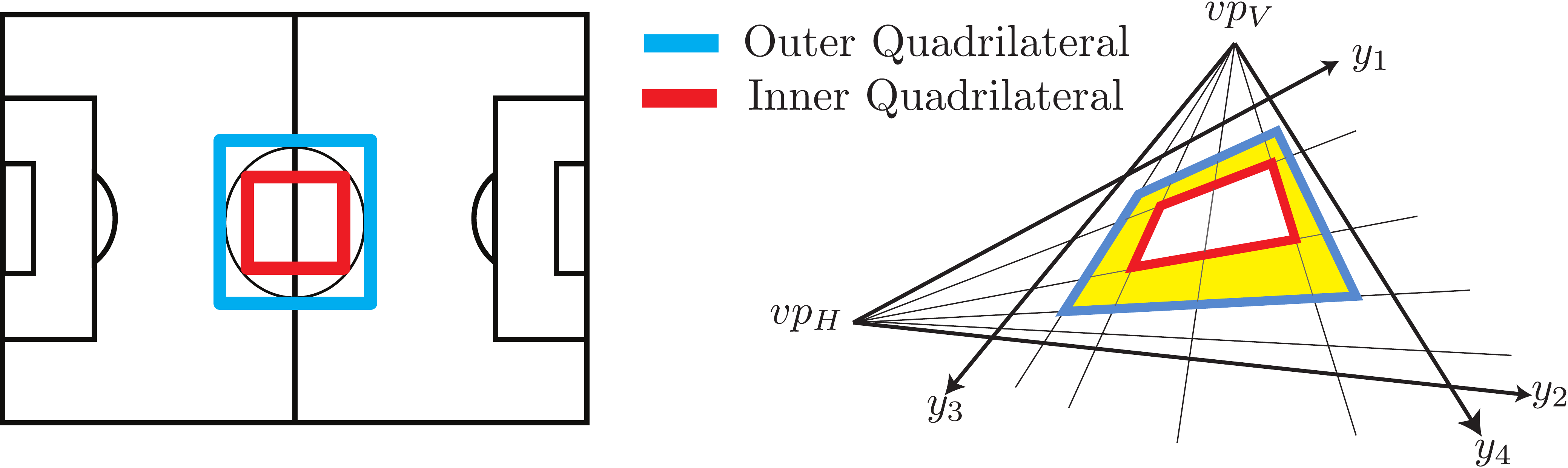}
\vspace{-0.6cm}
\caption{For each circle $C$ in the model, the projections of the inner (red) and outer (blue) quadrilaterals can be obtained using cross ratios. The potential $\phi_{C}(x,y)$ is the percentage of non-vp line pixels in the yellow region.}
\label{fig:circle-pot}
\end{figure}

A soccer field has white markings corresponding to a full circle centered at the middle of the field and two  circular arcs next to the penalty area,  all three with the same radius. 
When the geometric model of the field undergoes a homography $H$, these circular shapes transform to conics in the image. 
Similar to the line potentials, we seek to construct potential functions that count the percentage of supporting pixels for each circular shape given a hypothesis field $y$. These supporting pixels are edge pixels 
that do not fall on any line segments belonging to $vp_V$ or $vp_H$. 
Unlike the projected line segments, the projected circles are not aligned with  the grid $\Y$. 
However, as shown in Fig. \ref{fig:circle-pot}, we note that there are two unique inner and outer rectangles for each circular shape in the model which transform in the image $x$ to quadrilaterals aligned with the vanishing points. Their position in the grid can be computed similarly to lines using cross ratios. We define a potential $\phi_{C_i}(x,y)$ $i=1,2,3$ for each conic which simply counts the percentage of (non horizontal/vertical) line pixels inside the region defined by the two quadrilaterals.  

\vspace{-2mm}
\section{Exact Inference via Branch and Bound}

Note that the cardinality of our configuration space $\Y$, i.e. the number of hypothesis fields, is of the order $O(N_H^2N_V^2)$, which  is a very large number. In this section, we show how   to solve the inference task  in Eq.~\eqref{eq:inference} efficiently and exactly. Towards this goal, we design a    branch and bound  \cite{Lampert2009} (BBound) optimization over the space $\Y$ of all parametrized soccer fields. 
We take advantage of generalizations of integral images to 3D \cite{Schwing2012a} to compute our bounds very efficiently. 

Our BBound  algorithm thus requires three key ingredients:
\begin{enumerate}
\item A branching mechanism that can  divide any set into two disjoint subsets  of parametrized fields.
\item A set function $\bar{f}$ such that $\bar{f}(Y) \geq \max_{y\in Y} w^t\phi(x, y)$.
\item A priority queue which orders sets of parametrized fields $Y$ according to $\bar{f}$. 
\end{enumerate}

In what follows, we describe the first two components in detail.

\vspace{-2mm}
\subsection{Branching}
Suppose that $Y = \prod_{i=1}^4[y_{i,min}, y_{i,max}] \subset \Y$ is a set of hypothesis fields. At each iteration of the branch and bound algorithm we need to divide $Y$ into two disjoint subsets $Y_1$ and $Y_2$ of hypothesis fields. This is achieved by dividing the largest interval $[y_{i,min}, y_{i,max}]$ in half and keeping the other intervals the same. 

\begin{figure}[t]
\vspace{-0.6cm}
     \centering
\subfloat[]{
	\includegraphics[width=0.5\linewidth]{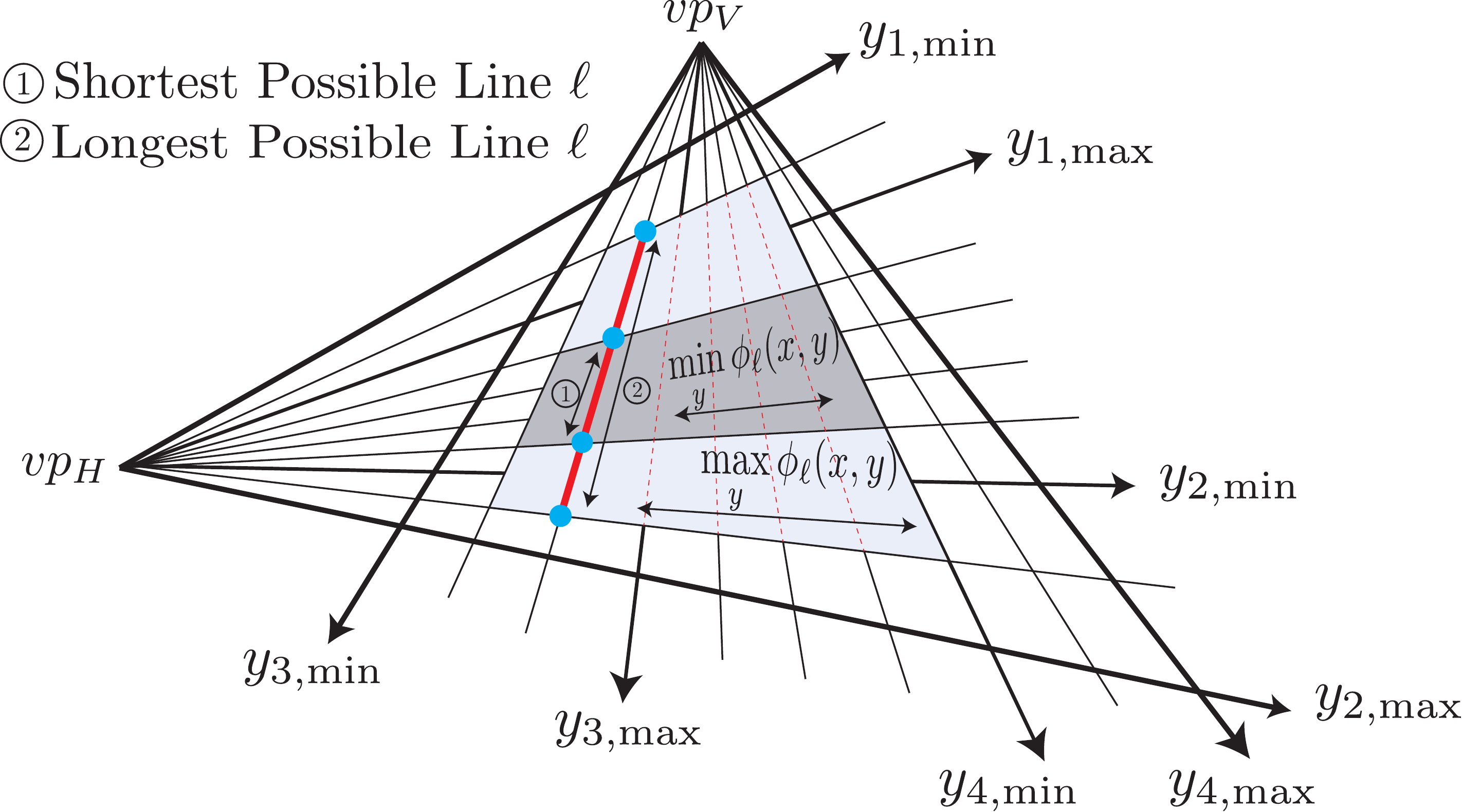}
	\label{fig:line-bound}
	 }
\subfloat[]{
	\includegraphics[width=0.5\linewidth]{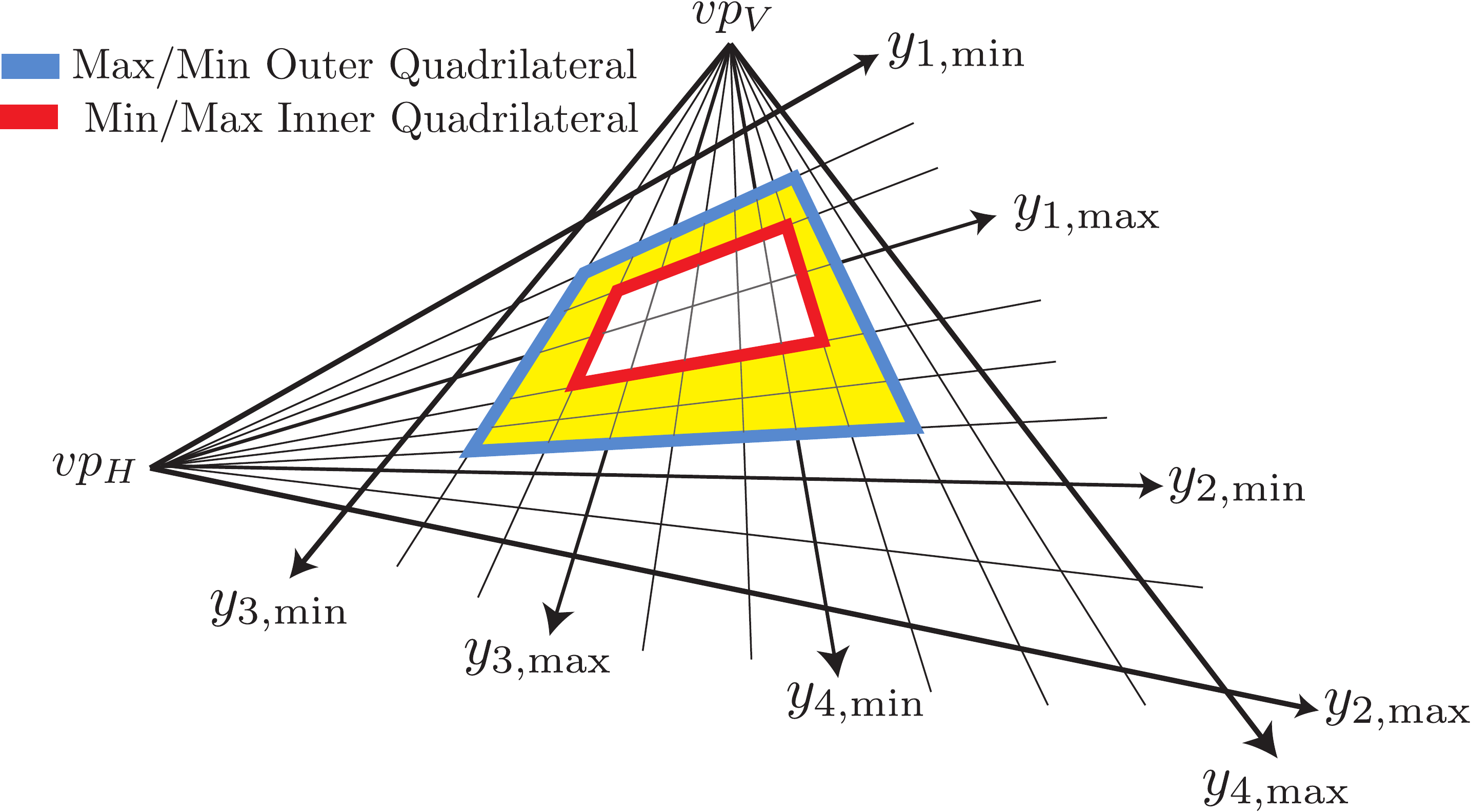}
	\label{fig:circle-bound}
  }
     \label{fig:bounding}
     \vspace{-4mm}
     \caption{(a) Finding the lower and upper bounds for a line correspond respectively to the $\min$ and $\max$ operations. (b) The upper/lower bound for $\phi_{C_i}(x,y)$ is the percentage of non-vp line pixels in the yellow region which is restricted by the max/min outer quadrilateral and the min/max inner quadrilateral.}
     \vspace{-1mm}
\end{figure}

\vspace{-2mm}
\subsection{Bounding}

We need to construct a set function $\bar{f}$ that upper bounds  $w^T\phi(x,y)$ for all $y \in Y$ where $Y \subset \Y$ is any  subset of parametrized fields. Since all potential function components of $\phi(x,y)$ are positive proportions, we decompose $\phi(x,y)$ into potential with strictly positive weights and those with weights that are either zero or negative: 
\begin{align}
	w^T\phi(x,y) = w_{neg}^T\phi_{neg}(x,y)+ w_{pos}^T\phi_{pos}(x,y)
	\label{eq:all}
\end{align}
with $w_{neg}$, $w_{pos}$  the vector of negative and positive weights respectively. 

We define the upper bound on Eq. \eqref{eq:all} to be the sum of an upper bounds on the positive features and a lower bound on the negative ones, 
\begin{align}
	\bar{f}(Y) = w_{neg}^T\bar{\phi}^{neg}(x,Y)+ w_{pos}^T\bar{\phi}^{pos}(x,Y)
\end{align}
It is trivial to see that this is a valid bound. 
In what follows, we construct a lower bound and an upper bound for all the potential functions of our energy. 

\vspace{-2mm}
\subsubsection{Bounds for the Grass Potential:} 
Let  $y_{\cap} := (y_{1,max}, y_{2,min}, y_{3,max}, y_{4,min})$ be the smallest possible field in  $Y$, and let  $y_{\cup} := (y_{1,min}, y_{2,max}, y_{3,min}, y_{4,max})$ be the largest. 
We now show how to construct the bounds for  $\phi_{grass-in}(x,y)$, and note that one can construct the other grass potential bounds in a similar fashion. Recall that $\phi_{grass-in}(x,y)$  counts the percentage of grass pixels inside the field. Since any possible field $y \in Y$ is contained within the smallest and largest possible fields $y_{\cap}$ and $y_{\cup}$ (Fig. \ref{fig:grass_image}b), we can define the the upper bound as the percentage of grass pixels inside the largest possible field and the lower bound as the percentage of grass pixels inside the smallest possible field. Thus: 
\[
	\bar{\phi}_{grass-in}^{pos}(x,Y) = \phi_{grass-in}(x,y_{\cap}),  \quad
	\bar{\phi}_{grass-in}^{neg}(x,Y) = \phi_{grass-in}(x,y_{\cup})
\]
We refer the reader to Fig. \ref{fig:grass_image}(b) for an illustration.



\vspace{-3mm}
\subsubsection{Bounds for the Line Potentials:}
We compute our bounds by finding a lower bound and an upper bound for each line independently. Since the method is similar for all the lines, we will illustrate it only for the left vertical penalty line $\ell$ of (Fig. \ref{fig:cross-ratios}a). For a hypothesis set of fields $Y$, we find the upper bound $\bar{\phi}_{\ell}^{pos}(x,Y)$ by computing the maximum value of $\phi_{\ell}(x,y)$ in the horizontal direction (i.e. along the rays from $vp_V$) but only for the maximal extended projection of $\ell$ in the vertical direction (i.e. along the rays from $vp_H$). This is demonstrated in (Fig. \ref{fig:line-bound}). Finding a lower bound consists instead of finding the minimum $\phi_{\ell}(x,y)$ for minimally extended projections of $\ell$.

\hide{
Note that for any $y \in Y$, the ray segment $r$ from $vp_V$ on which $\ell'$ falls upon depends on the goalline rays $y_3$ and $y_4$. \raquel{why do you need $r$? why complicated notation with $q$ also?}
Also, the projection $\ell'$ on $r$ is restricted by rays $q_1$ and $q_2$ from $vp_H$ which depend on $y_1$ and $y_2$. 
For any combination of $y_1 \in [y_{1,min}, y_{1,max}]$ and $y_2 \in [y_{2,min}, y_{2,max}]$ we have a ray $r \in \set{r_1, \dots, r_{n_\ell}}$. 
The most left ray $r_1$ belongs to the fields restricted by touchlines $y_{3,min}$ and $y_{4,min}$ and the most right ray $r_{n_{\ell}}$ belongs to the fields restricted by $y_{3,max}$ and $y_{4,max}$. The horizontal restrictions of $\ell'$ can be found in a similar fashion by computing the upper and lower bounds of the rays $q_1$ and $q_2$. This is depicted in \todo{(Ref to fig)}. 

We define the upper bound function $\bar{\phi}^{pos}_{\ell}$ for the line $\ell$ to be the maximum value $\phi_{\ell}$ can take on the rays $r_1$ to $r_{n_{\ell}}$ and restricted below and above by $q_{1,\min}$ and $q_{2,\max}$ respectively as depicted in figure \todo{(Ref to fig)}. A lower bound is found in a similar fashion. 
}

Note that for a set of hypothesis fields $Y$, this task requires a linear search over all the possible rays in the horizontal (for vertical lines) at each iteration of branch and bound. However, as the branch and bound continues, the search space becomes smaller and finding the maximum becomes faster.

\vspace{-3mm}
\subsubsection{Bounds for the Circle Potentials:}
Referring back to the definition of the ellipse potentials $\phi_{C_i}(x,y)$ provided in section~\ref{sec:circle} and a set of hypothesis fields $Y$, we aim to construct lower and upper bounds for each circle potential. For an upper bound, we simply let $\phi_{C_i}^{pos}(x,Y)$ be the percentage of non-vp line pixels contained in the region between the smallest inner and largest outer quadrilaterals as depicted in (Fig. \ref{fig:circle-bound}). A lower bound is obtained in a similar fashion.

\subsection{Integral Accumulators for Efficient Potentials and Bounds}



We construct five 2D accumulators corresponding to the grass pixels, non-grass pixels, horizontal line edges, vertical  line edges, and non-vp line edges. In contrast to \cite{Viola2001}, and in the same spirit of \cite{Schwing2012a}, our accumulators are aligned with the two orthogonal vanishing points and count the fraction of features in the regions of $x$ corresponding to quadrilaterals restricted by two rays from each vanishing point. 
In this manner, the computation of a potential function over any region in $\Y$ boils down to four accumulator lookups. Since we defined all the lower and upper bounds in terms of their corresponding potential functions, we use  the same accumulators to compute the bounds in constant time. 


\vspace{-3mm}
\subsection{Learning}

We use structured support vector machine (SSVM) to learn the parameters $w$ of the log linear model. 
Given a  dataset composed of  training pairs $\set{x^{(n)},y^{(n)}}_{i=1}^N$, we obtain $w$ by minimizing the following objective 
\begin{align}
	\min_{w}\frac{1}{2}\norm{w}^2 + \frac{C}{N}\sum_{n=1}^N \max_{y \in \Y}\big( \Delta(y^{(n)},\hat{y})+ w^T\phi(x^{(n)},\hat{y}) - w^T\phi(x^{(n)},y^{(n)}) \big) \label{eq:learning}
\end{align}
where $C>0$ is a regularization parameter and $\Delta:\Y \times \Y \to \R^{+}\cup\set{0}$ is a loss function measuring the distance between the ground truth labeling $y^{(n)}$ and a prediction $\hat{y}$,  with $\Delta(y^{(n)},y) = 0$ if and only if $y = y^{(n)}$. In particular, we employ the parallel cutting plane implementation of \cite{Schwing2013}. 

The loss function is defined very similarly to  $\phi_{grass-in}(x,y)$. However, instead of segmenting the image $x^{(n)}$ to grass vs. non-grass pixels,   we segment the grid $\Y$ to field vs. non-field cells by reprojecting the ground truth field into the image. 
Then given a hypothesis field $y$, we define the loss for a training instance $(x^{(n)}, y^{(n)})$ to be 
\[
 	\Delta(y^{(n)},y) = 1 -\frac{\big(\text{\# of field cells in $F_y$}\big) + \big(\text{\# of non-field cells outside of $F_y$ }\big) }{\text{Number of cells in $\Y$ }}
\]
Note that the loss can be computed using integral accumulators, and loss augmented inference can be performed efficiently and exactly using our BBound.

\vspace{-2mm}
\section{Vanishing Point Estimation}
\label{sec:VP}

In a Manhattan world, such as a soccer stadium, there are three principal orthogonal vanishing points. Our goal is the find the two orthogonal vanishing points $vp_V$ and $vp_H$ that correspond to the lines of the soccer field. We forgo the estimation of the third orthogonal vanishing point since in a broadcast image of the field there are not usually many lines corresponding to this vanishing point. However, a reasonable assumption is to take the direction of the third vanishing point to be in the direction of gravity since the main camera rarely rotates. 
We find an initial estimate of the positions of $vp_H$ and $vp_V$  by deploying the line voting procedure of \cite{Hedau2009}. 
This procedure is robust when there are sufficiently enough line segments for each vanishing point. In some cases, for example when the camera is facing the centre of the field (Fig. \ref{fig:vp_grass}b), there might not be enough line segments belonging to $vp_V$ to estimate its position reliably but enough to distinguish its corresponding line segments. In this case, we take the line segments that belong to neither vanishing point and fit an ellipse \cite{Fitzgibbon1999} which is an approximation to the conic in the centre of the field. We then take the 4 endpoints of the ellipses' axes and also one additional point corresponding to the crossing of the ellipses' minor axis from the grass region to non-grass region to find an approximate homography which in turn gives us an approximate $vp_V$.

\vspace{-2mm}
\section{Experiments}

For assessing our method, we recorded 12 games from the World Cup 2014 held in Brazil. Out of these games we annotated 259 images with the ground truth fields and also the grass segmentations. We used 6 games with 154 images for training and validation sets, and 105 images from 6 other games for the test set. The images consist of different views of the field with different grass textures. Some images, due to the rain,  seem blurry and lack some lines. We remind the reader that these images do not have a temporal ordering. 

Out of the 259 images, the vanishing point estimation failed for 5 images in the training/validation set and for 3 images in the test set. We discarded these failure cases from our training and evaluation. 
In what follows we assess different components of our method.

\vspace{-2mm}
\paragraph{\bf Grass Segmentation:} is a major component of our method since it has its own potentials and is also used for restricting the set of detected line segments in the image to the ones that correspond to white markings of the field. 
Most of the existing approaches mentioned in the related work's section, use heuristics based on color and hue information to segment the image into grass vs. non-grass pixels. We found these heuristics to be unreliable at times since the texture and color of the grass can be different from one stadium to another. Moreover, at some games, the spectators wear clothing with similar colors to the grass which further makes the task of grass segmentation difficult. 

As a result, we fine-tune the CNN component of the DeepLab network \cite{chen14semantic} on the train/validation images annotated with grass and non-grass pixels.
Our trained CNN grass segmentation method achieves an Intersection over Union (IOU) score of 0.98 on the test set. Some grass segmentation examples are shown in Fig.~\ref{fig:vp_grass}.


\vspace{-2mm}
\paragraph{\bf Ablation studies:}
In Table~\ref{table:ablation} we present the IOU score of test images based on employing different potentials in our energy function. For each set of features we used the weights corresponding to the value of $C$ that maximizes the IOU score of the validation set. We notice that just including the grass potentials achieves a very low test IOU of 0.57. This is expected since grass potentials by themselves do not take into account the geometry of the field. However, when we include line and circle potentials, the test IOU increases by about 30\%.

\begin{table}[t]
\centering
\caption{{\bf G} correspond to 4 weights for each grass potential. {\bf L}: all the lines share the same weight. {\bf C}: all the circles share the same weight. {\bf VerL}: all the vertical lines share the same weight. {\bf VerH}: all the horizontal lines share the same weight.}
\label{table:ablation}
\vspace{1mm}
\addtolength{\tabcolsep}{6pt}
\begin{tabular}{|l|l|l|}
\hline
Potentials    & Mean Test IOU & Median Test IOU \\ \hhline{|=|=|=|}
G             & 0.57          & 0.56            \\ \hline
G+L           & 0.85          & 0.93            \\ \hline
L+E           & 0.88          & 0.94            \\ \hline
G+L+C         & 0.89          & 0.94            \\ \hline
G+VerL+HerL+C & 0.90          & 0.94            \\ \hline
\end{tabular}
\vspace{-2mm}
\end{table}


\vspace{-2mm}
\paragraph{\bf Comparison of Our Method to Two Baselines:}
There is currently no baseline in the literature for automatic field localization in the game of soccer. As such, we derive two baselines based on our segmentation and line segment detection methods. As the first baseline, for each test image we retrieve its nearest neighbour (NN) image from the training/validation sets based on the grass segmentation IOU and apply the homography of the training/val image on the test image. The second baseline is similar but instead of the NN based on grass, we retrieve based on the  distance transform computed from the edges \cite{meijster2002}. Note that these approaches could be considered similar to the keyframe initialization methods of~\cite{Gupta2011a,Okuma2004a,Dubrofsky2008,Hess2007}. In contrast to those papers, here we retrieve the closest homography from a set of different games.    

In Table \ref{table:baseline}, we compare the IOU of these baseline with our learned branch and bound inference method. We observe that if we use only the grass potentials, the baseline is similar to the NN with grass segmentation. Using NN with line segment detections improves the baseline. When we introduce potential functions based on lines, the IOU metric is increased by about 30\%. Our method with the best set of features outperform the baseline by about 34\%. The best set of features that achieve an IOU of 90\% have 4 weights for the grass potentials, one shared weight for the vertical lines, one shared weight for the horizontal lines, and similarly one shared weight for the circles. 
By releasing our dataset and the annotations, we hope that other baselines will be established.

\begin{table}[t]
\centering
\caption{Comparison of the branch and bound inference method with two baselines}
\label{table:baseline}
\begin{tabular}{|c|c|c|}
\hline
method                   & Mean Test IOU   & Median Test IOU \\ \hhline{|=|=|=|}
Nearest Neighb. based on grass segmentation    & 0.56 & 0.64   \\ \hline
Nearest Neighb. based on lines distance transform    & 0.59 & 0.66   \\ \hline
our method with just grass potentials & $ 0.57$ & $ 0.56$   \\ \hline
our method with line potentials & $\geq 0.85$ & $\geq 0.93$   \\ \hline
our method best features (G+VerL+HorL+C) & 0.90 & 0.94   \\ \hline
\end{tabular}
\end{table}

\begin{figure}[t!]
     \centering
\subfloat{
\includegraphics[width=0.33\linewidth]{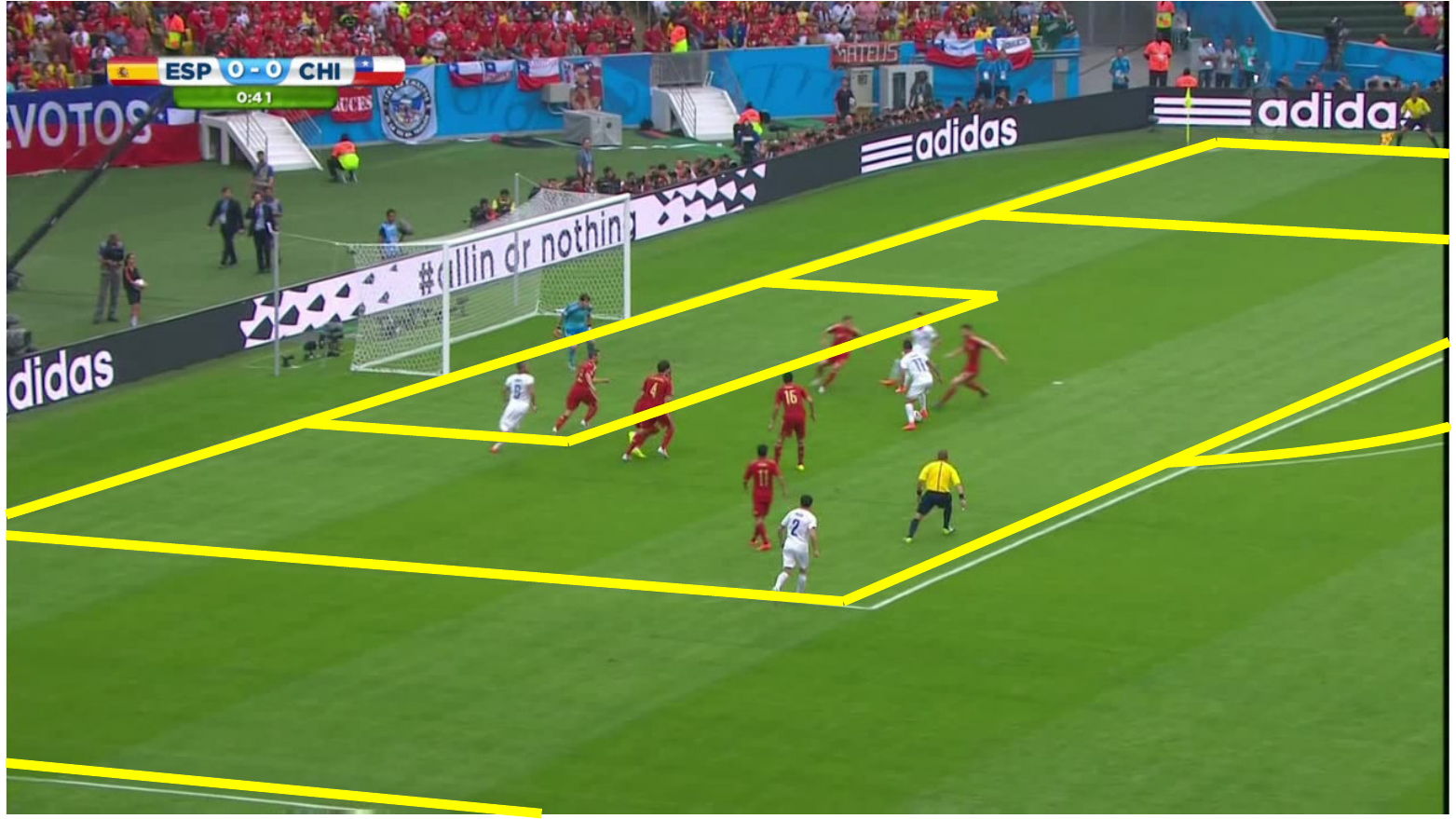}
	 }
\subfloat{
	\includegraphics[width=0.33\linewidth]{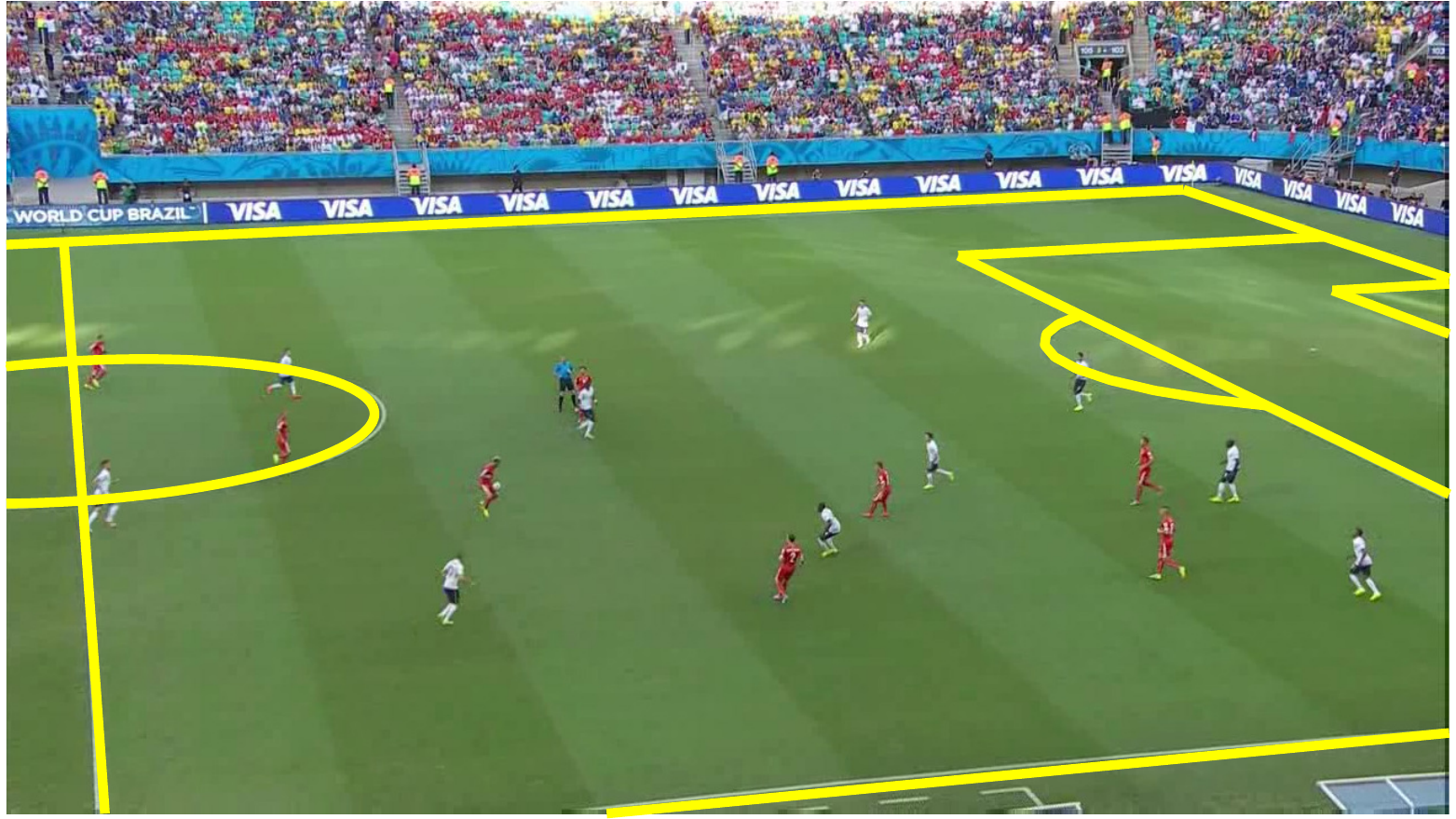}
  }
\subfloat{
       	\includegraphics[width=0.33\linewidth]{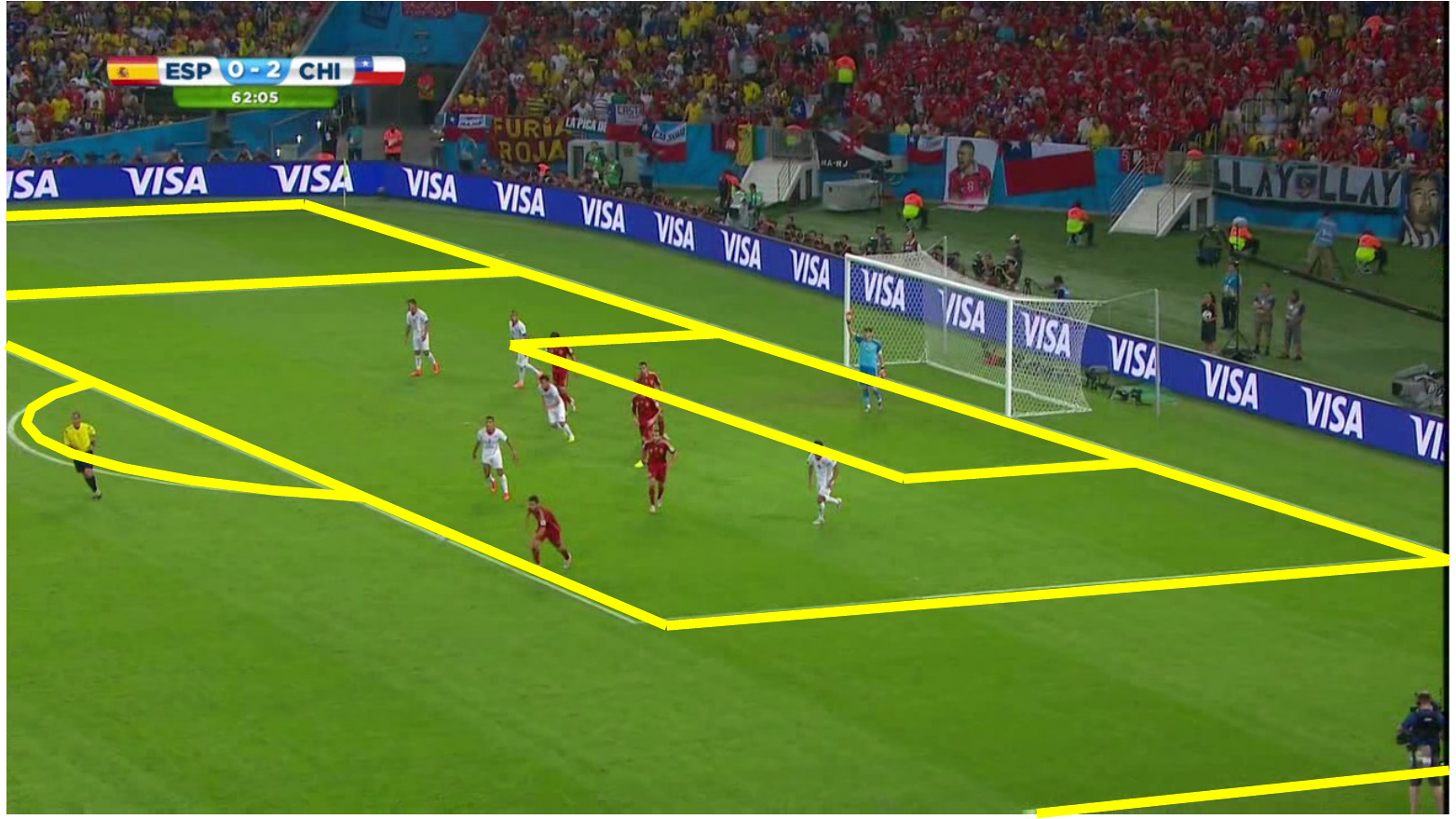}
  }
\\
\subfloat{
	\includegraphics[width=0.33\linewidth]{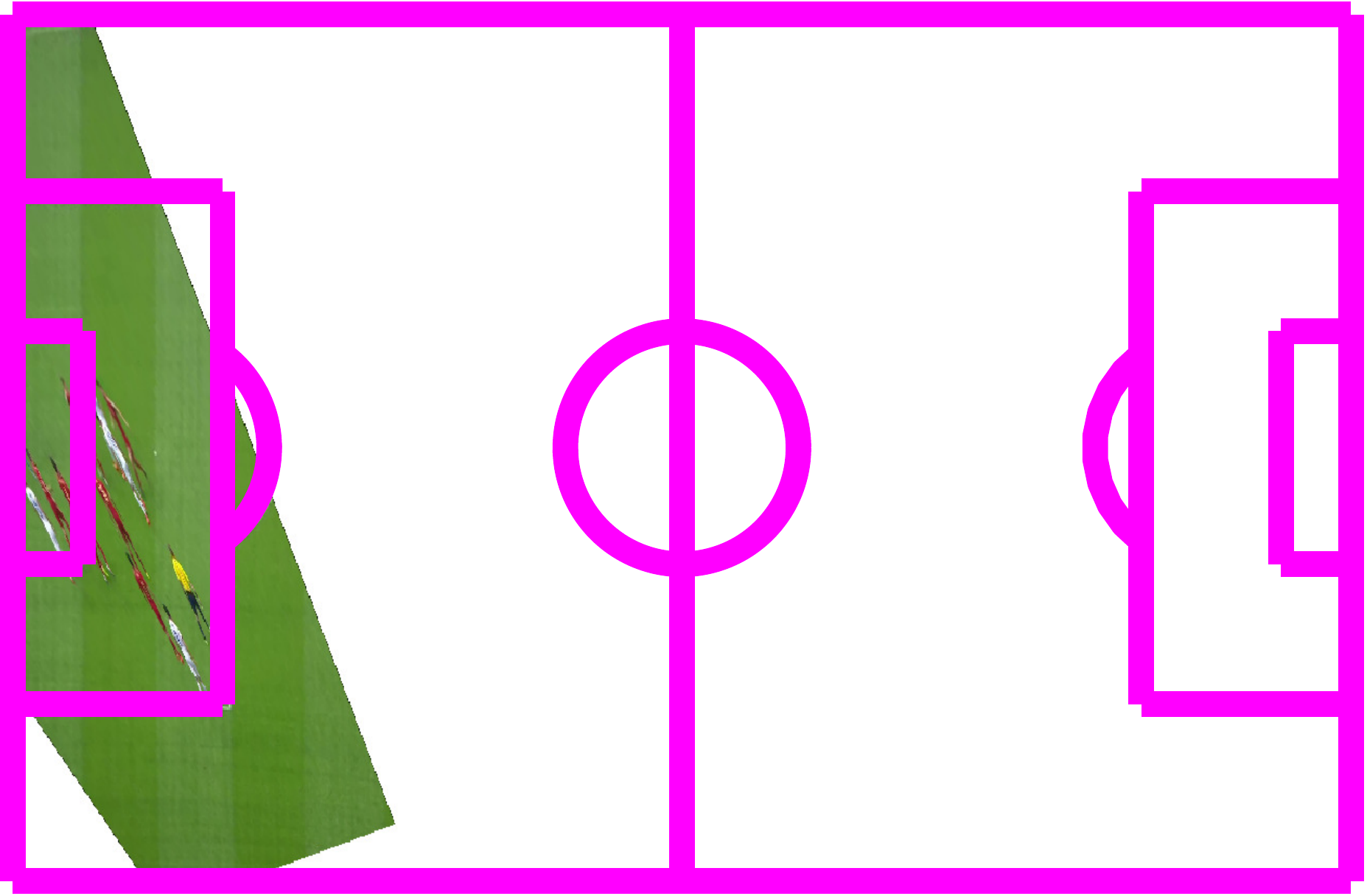}
	 }
\subfloat{
	\includegraphics[width=0.33\linewidth]{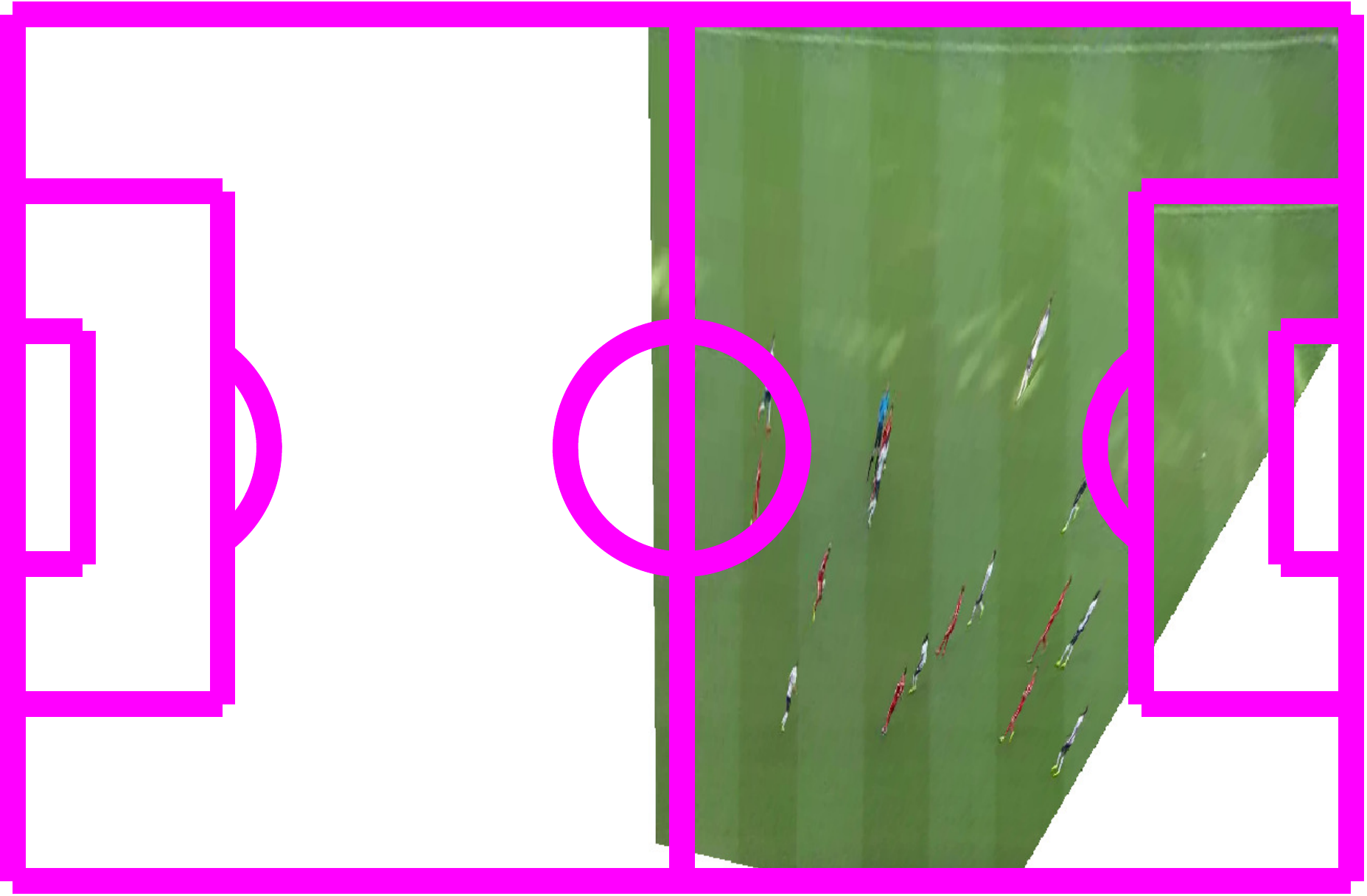}
  }
\subfloat{
       	\includegraphics[width=0.33\linewidth]{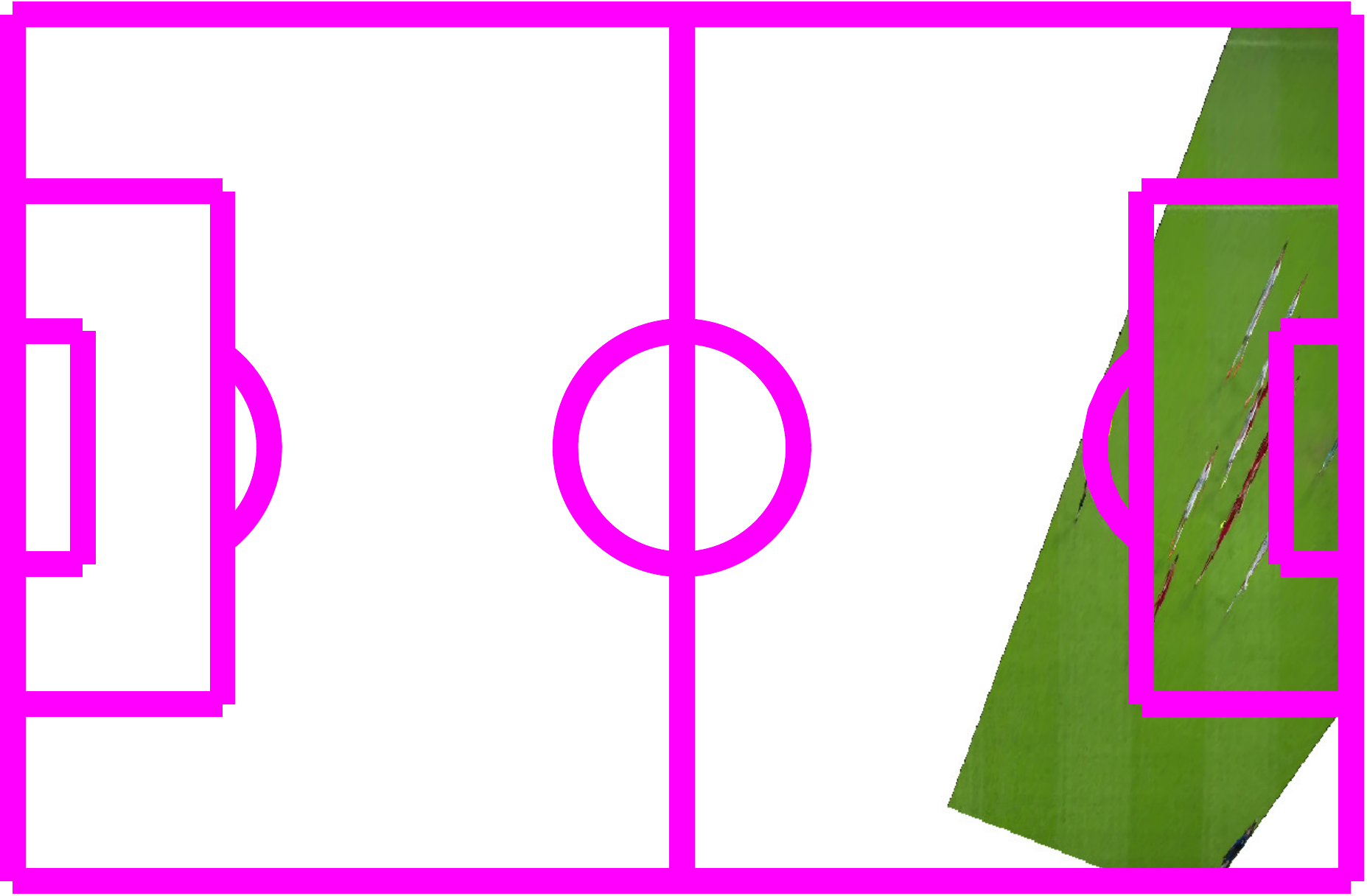}
  }
\\
\subfloat{
	\includegraphics[width=0.33\linewidth]{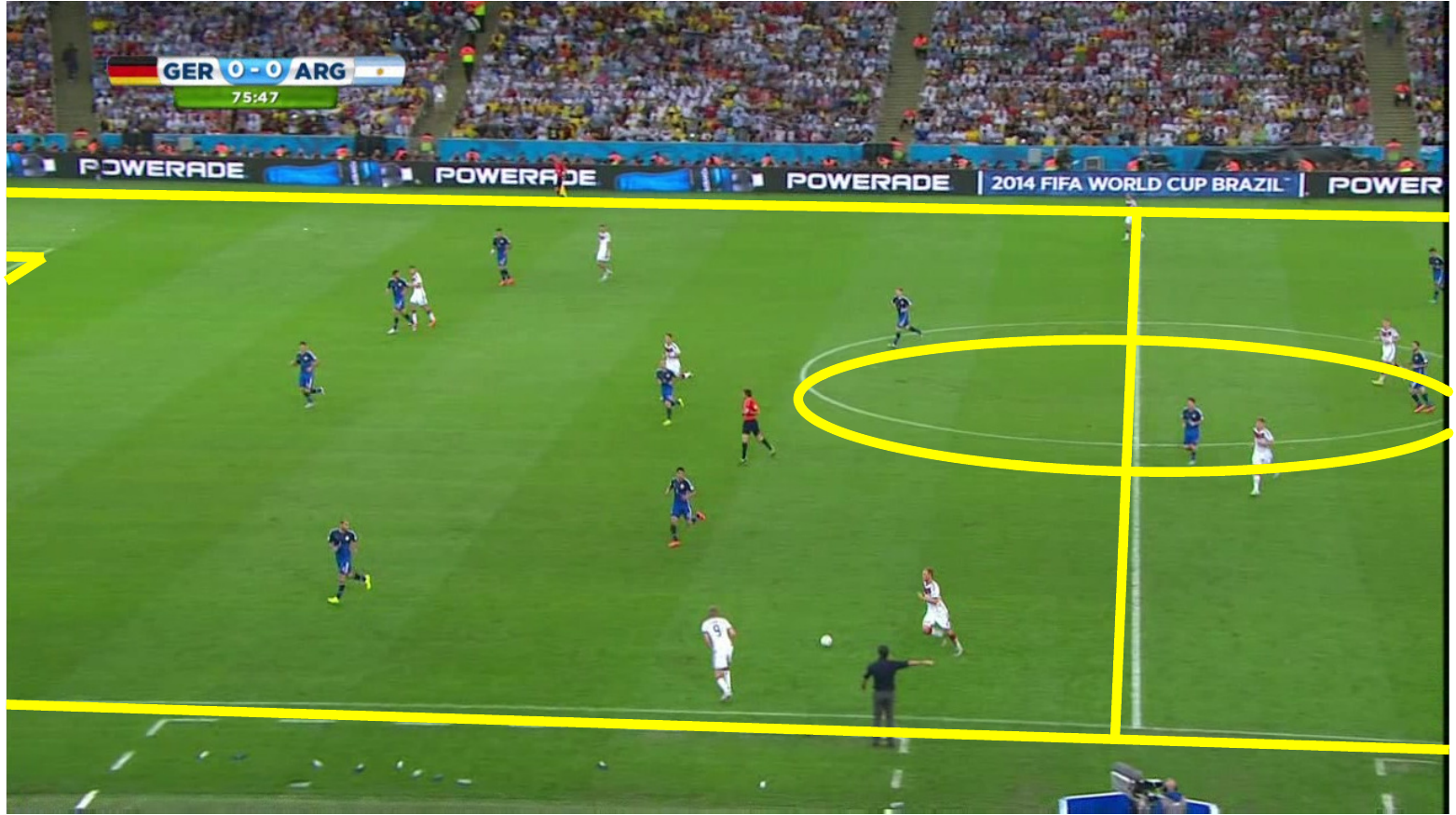}
	 }
\subfloat{
	\includegraphics[width=0.33\linewidth]{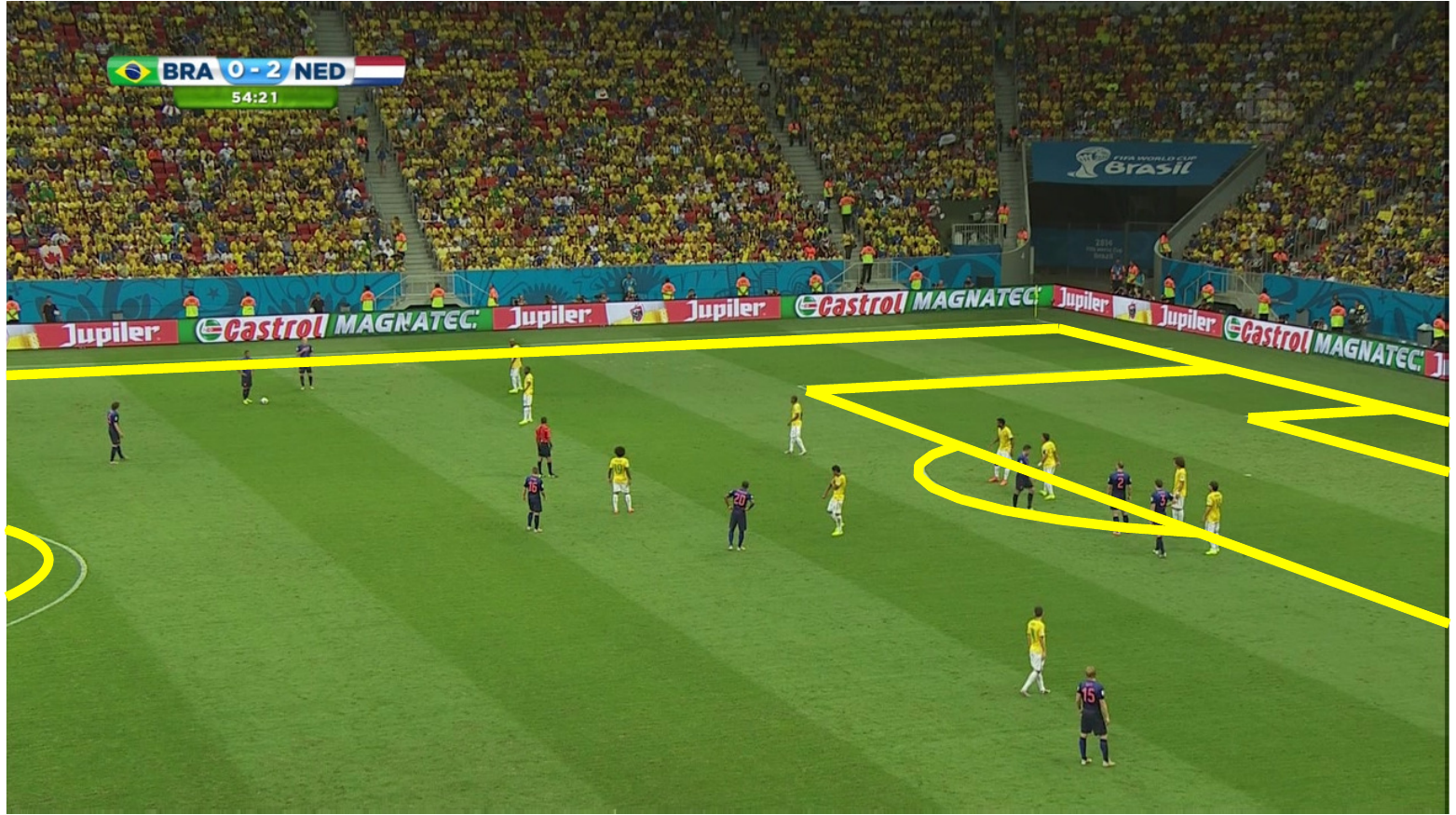}
  }
\subfloat{
       	\includegraphics[width=0.33\linewidth]{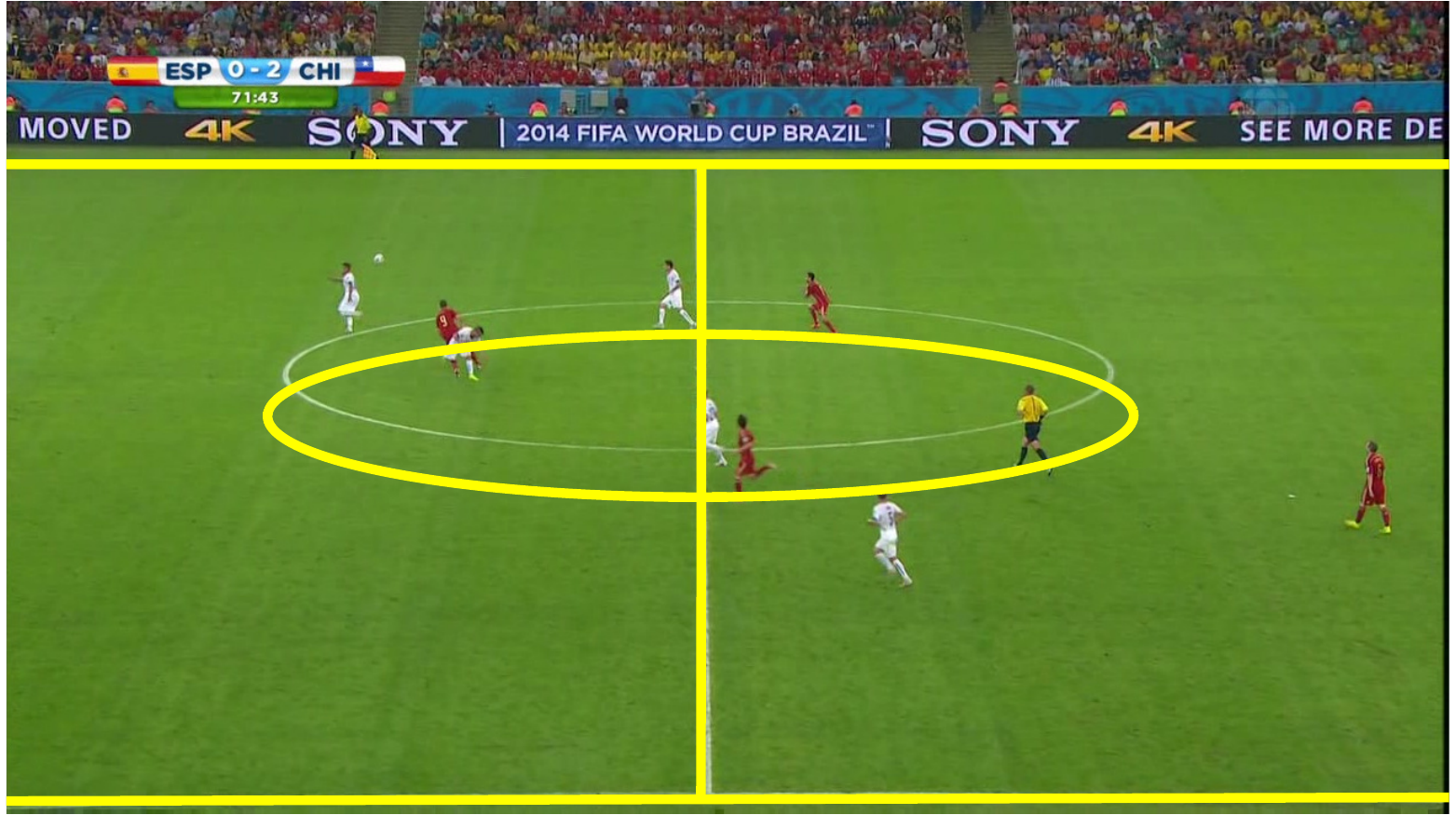}
  }
\\
\subfloat{
	\includegraphics[width=0.33\linewidth]{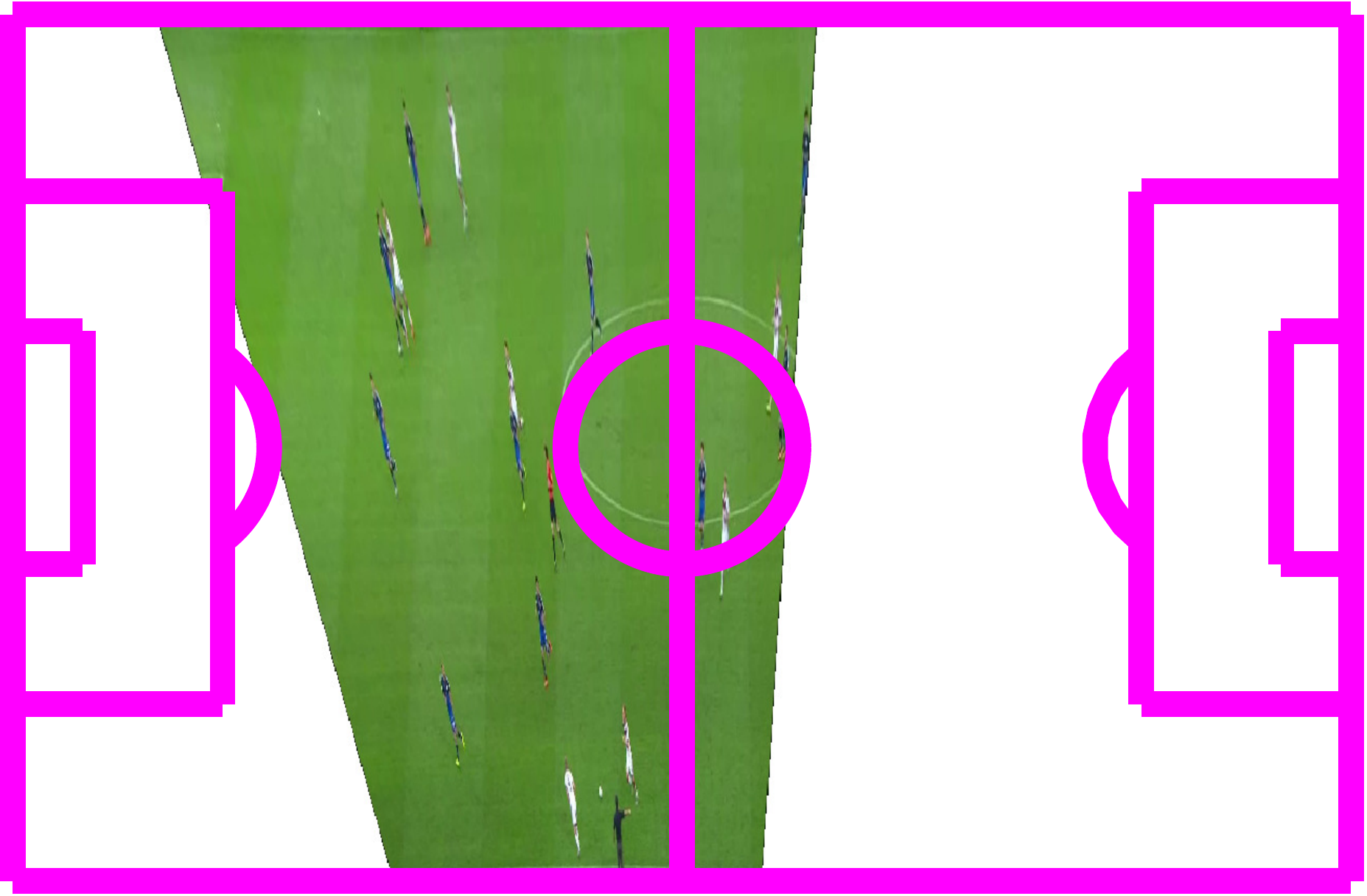}
	 }
\subfloat{
	\includegraphics[width=0.33\linewidth]{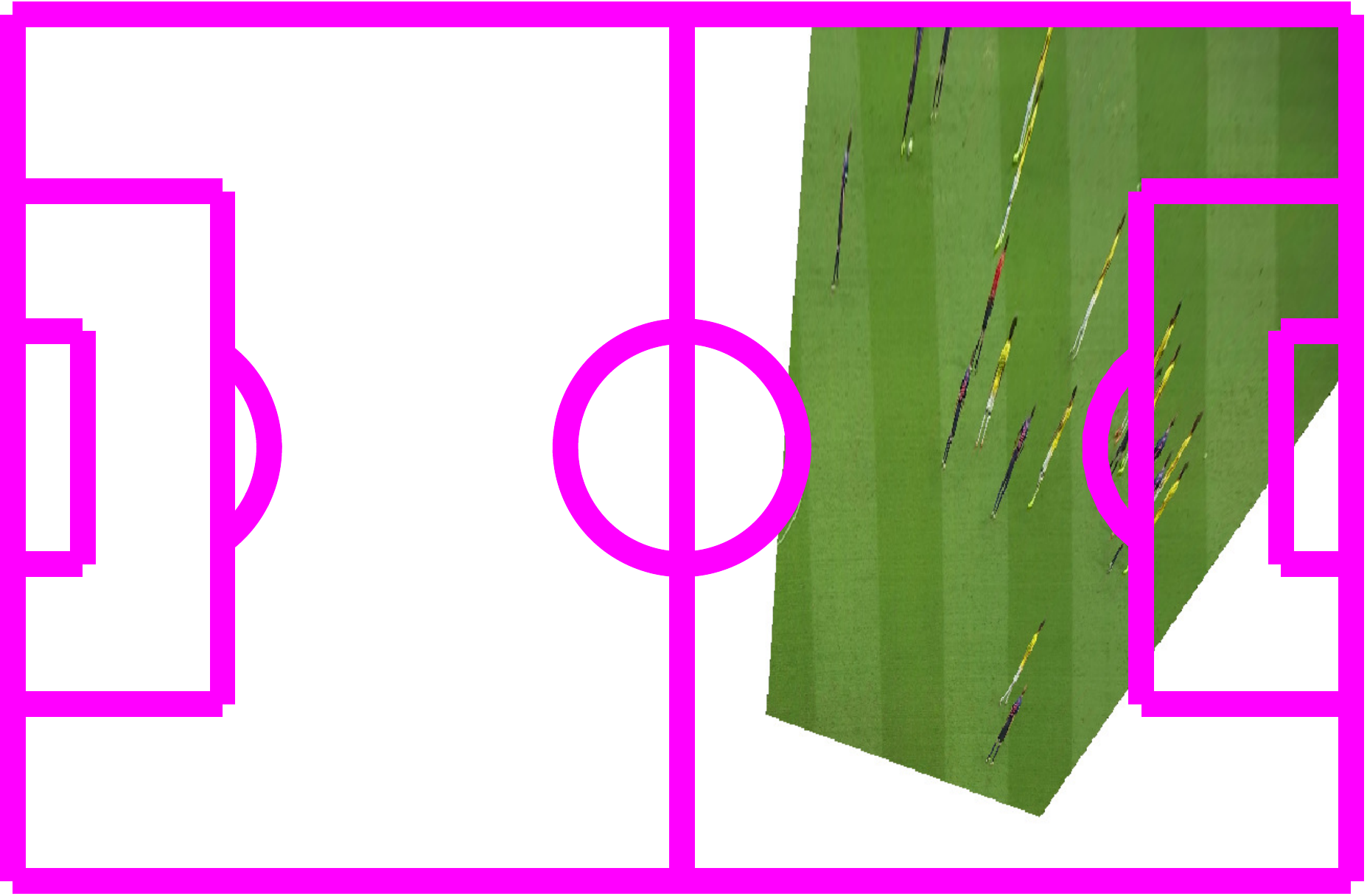}
  }
\subfloat{
       	\includegraphics[width=0.33\linewidth]{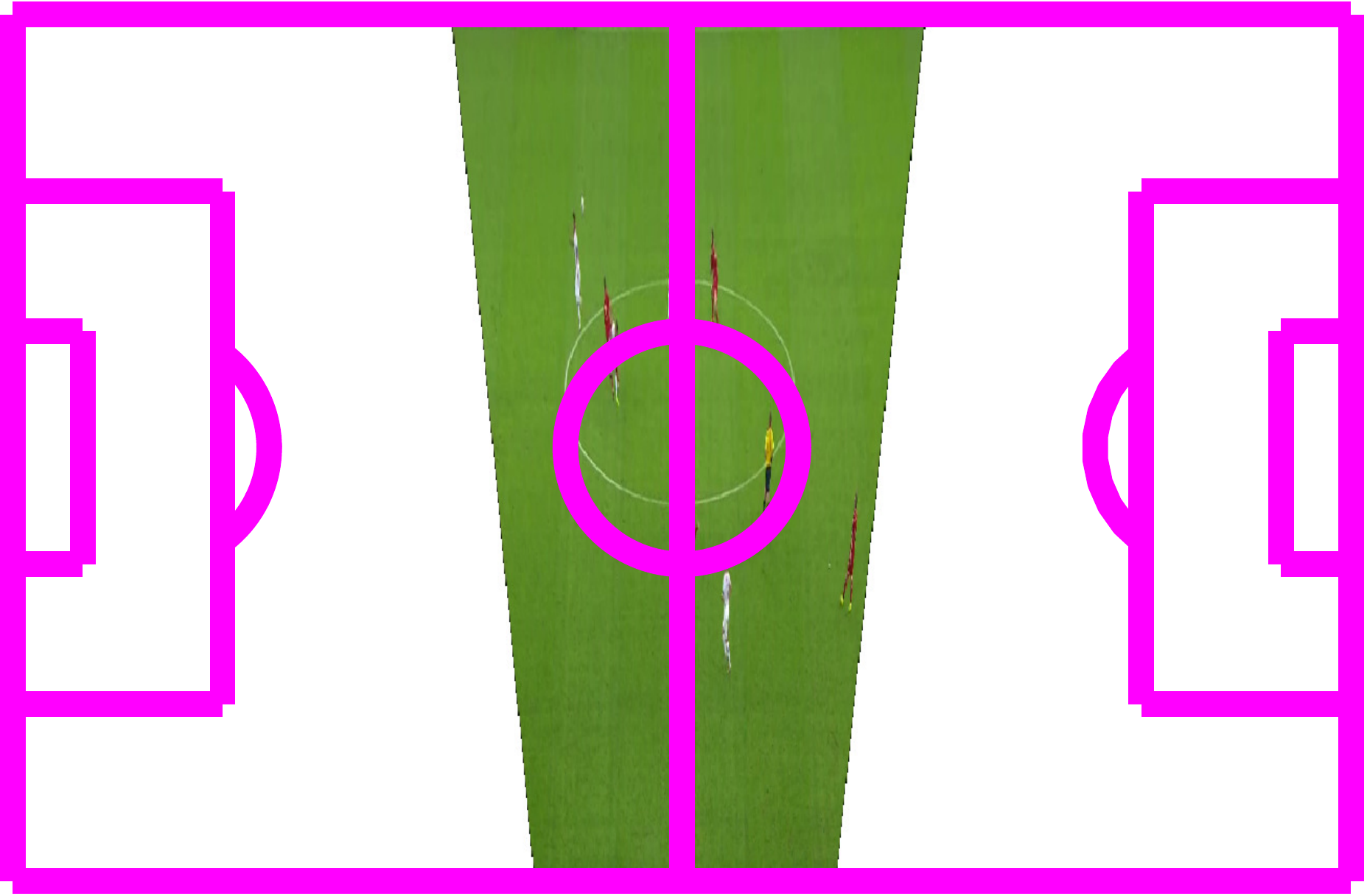}
  }\\
\subfloat{
	\includegraphics[width=0.33\linewidth]{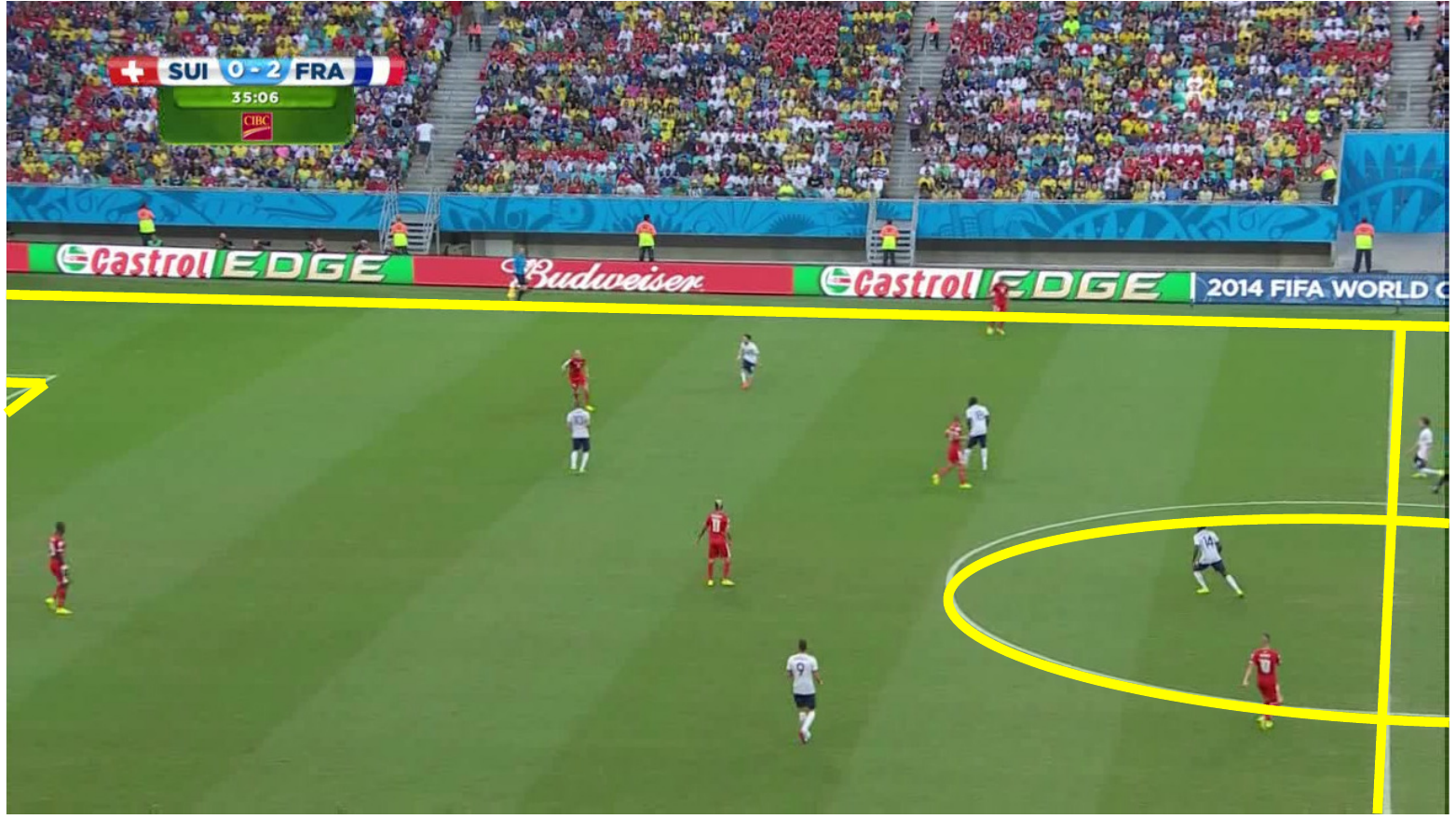}
	 }
\subfloat{
	\includegraphics[width=0.33\linewidth]{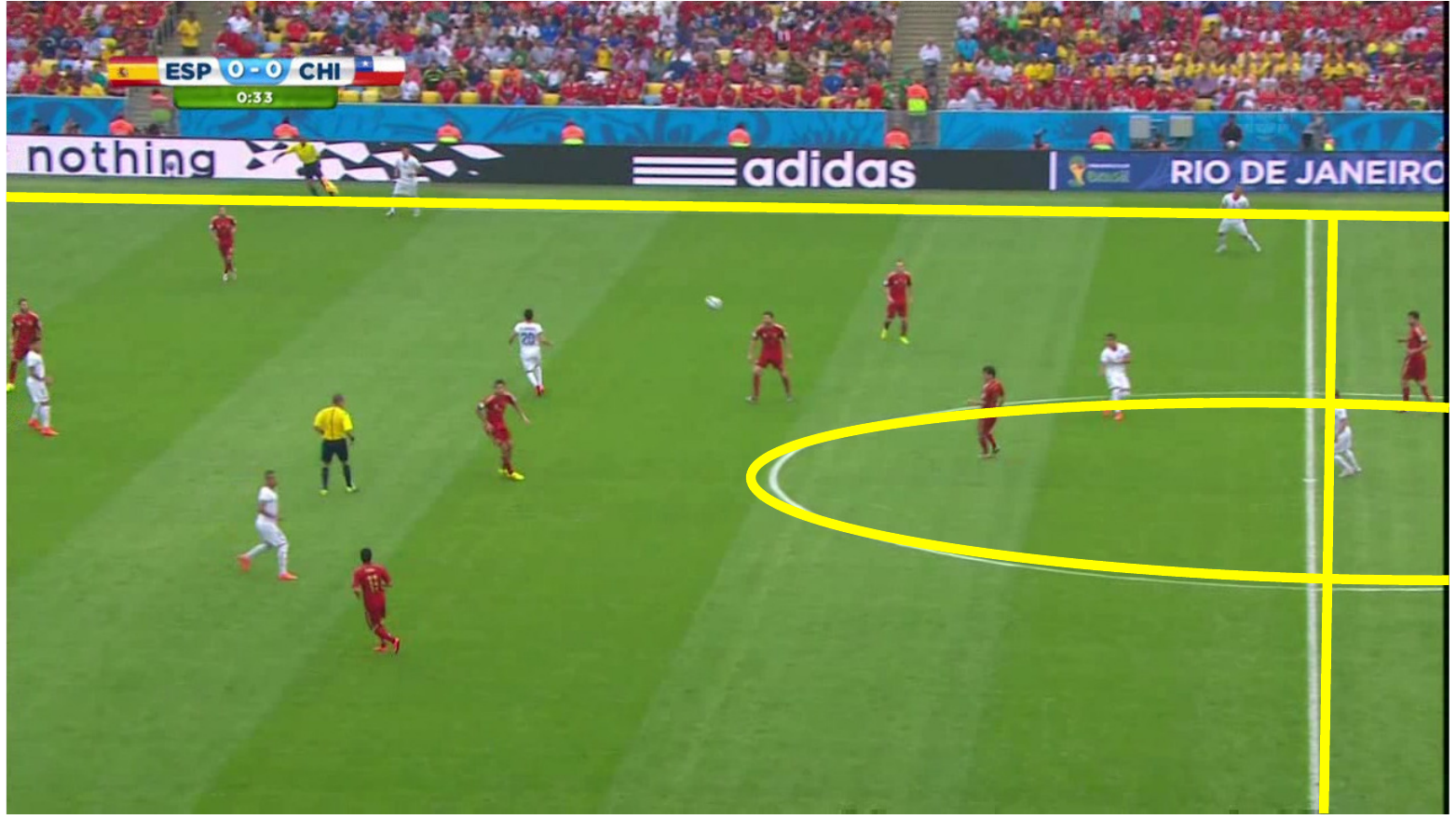}
  }
\subfloat{
       	\includegraphics[width=0.33\linewidth]{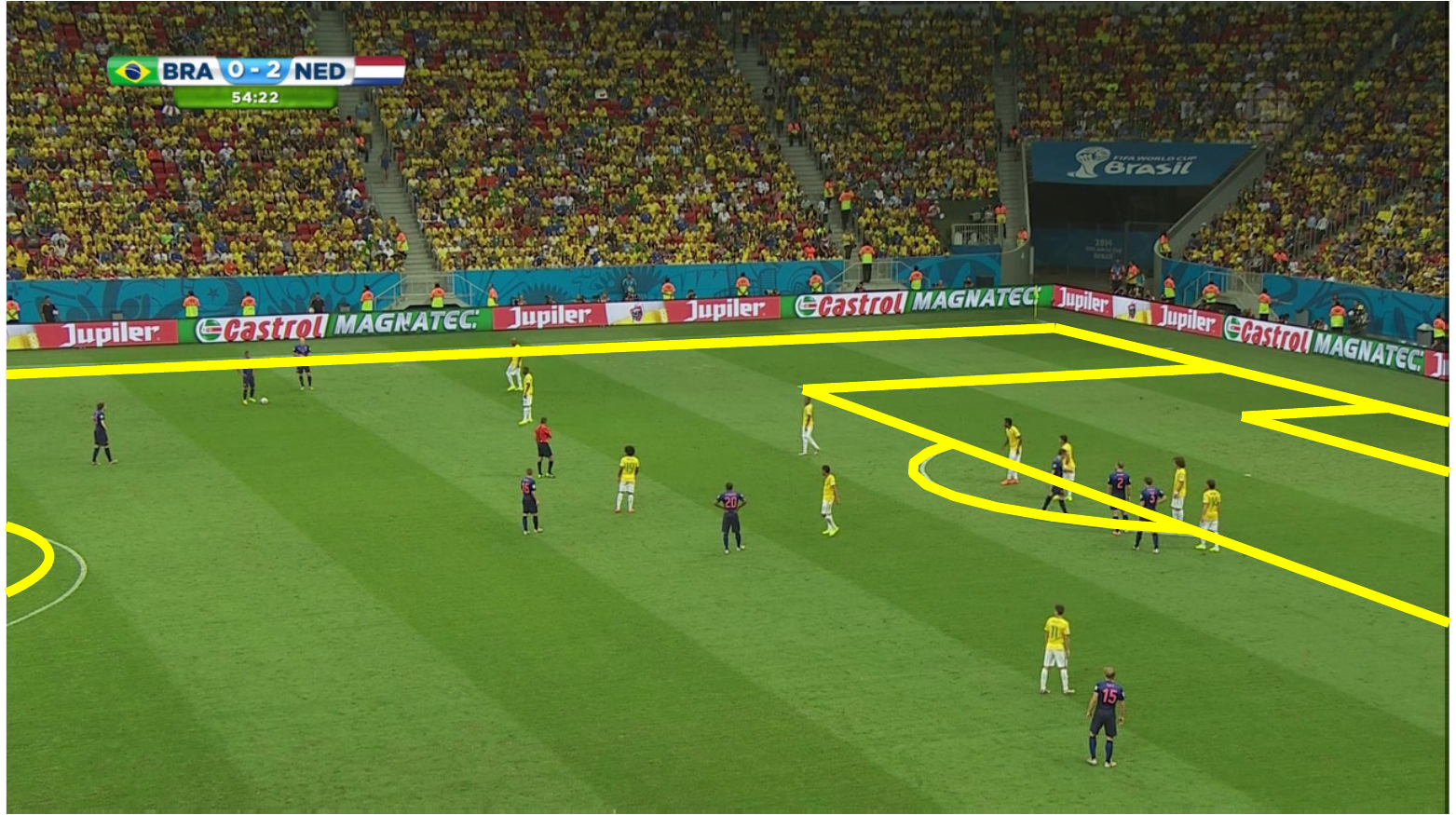}
  }
\\
\subfloat{
	\includegraphics[width=0.33\linewidth]{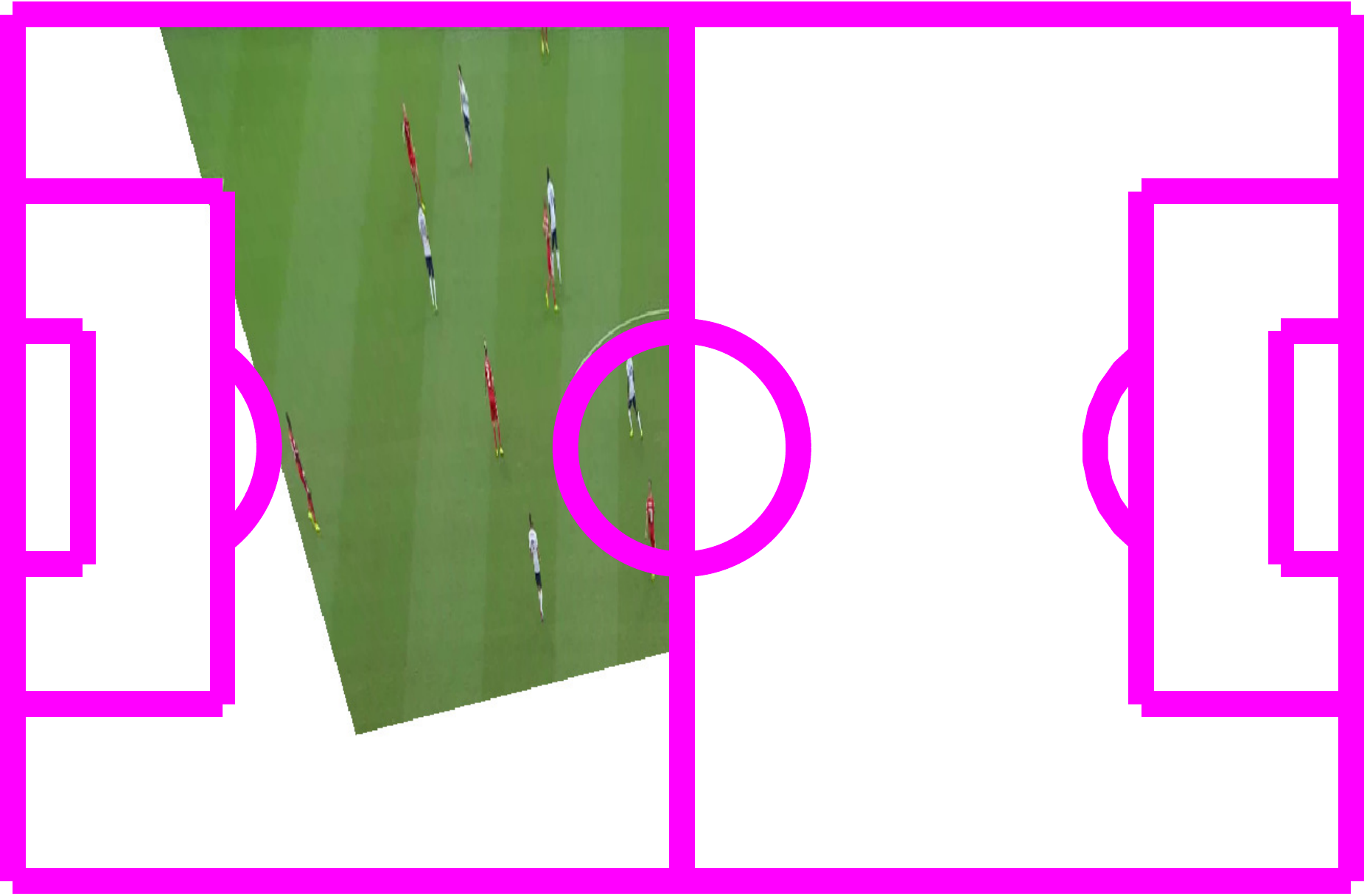}
	 }
\subfloat{
	\includegraphics[width=0.33\linewidth]{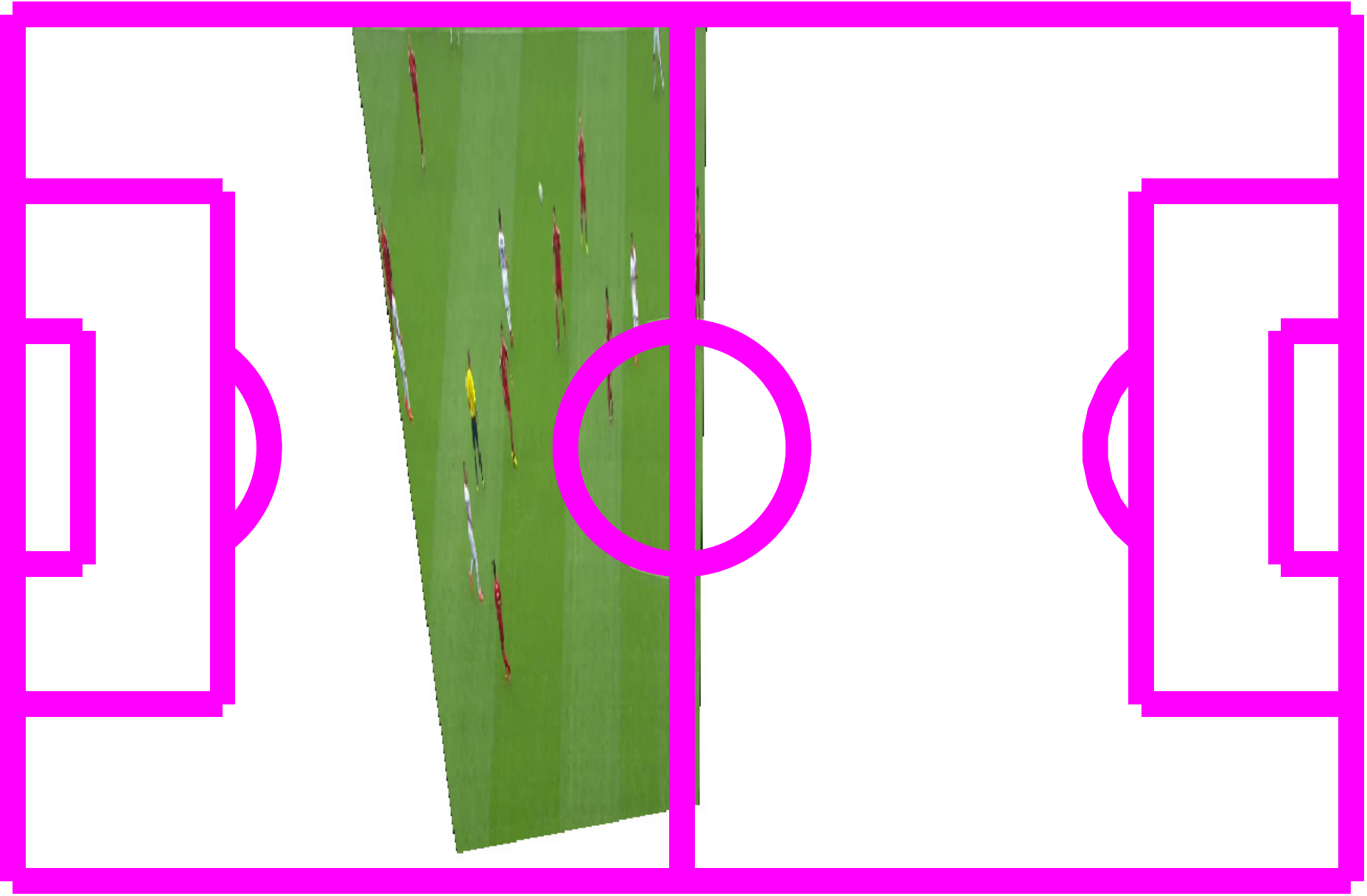}
  }
\subfloat{
       	\includegraphics[width=0.33\linewidth]{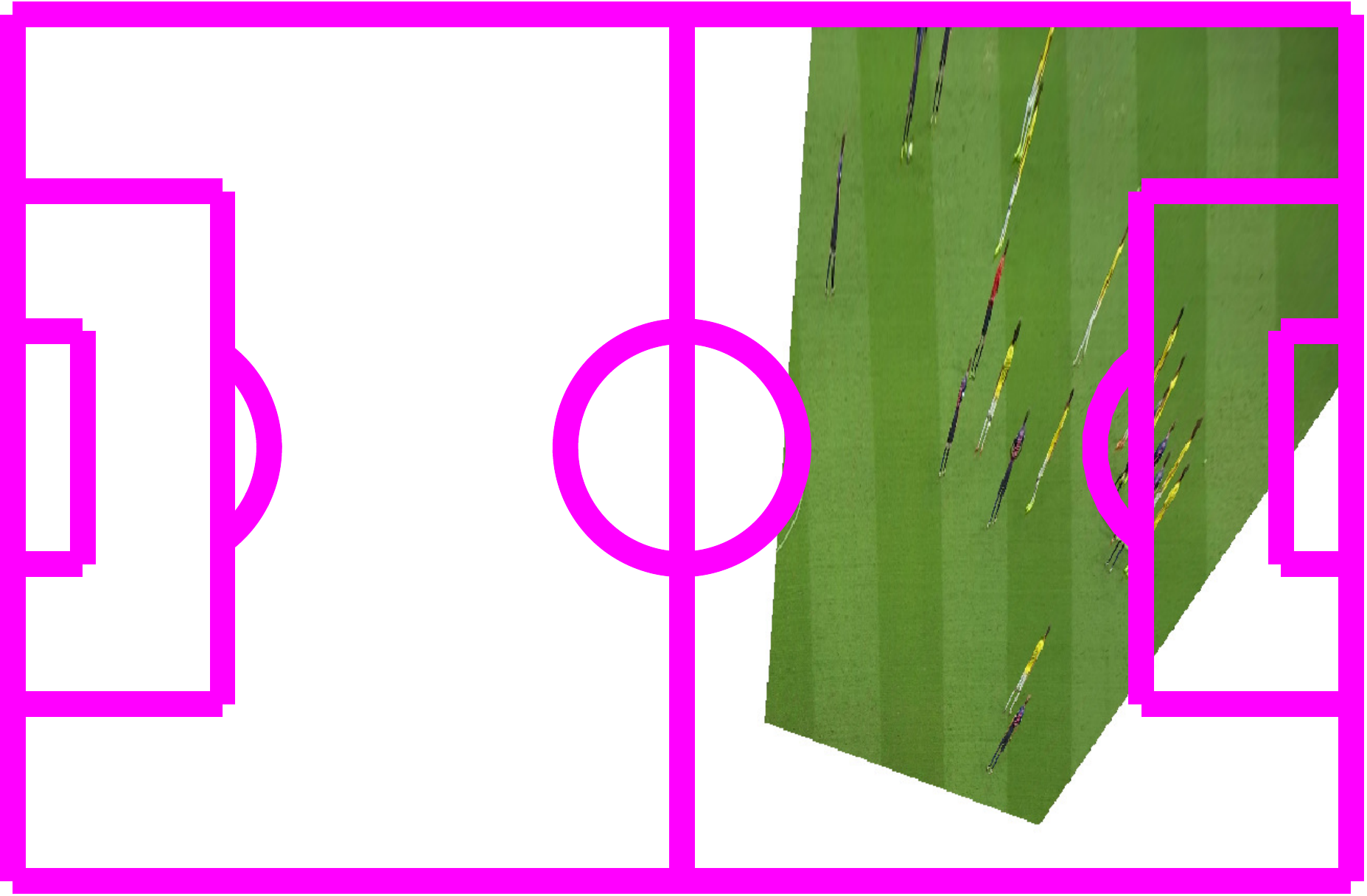}
  }

     \caption{Some examples of the obtained homography. The yellow lines correspond to the projection of the model lines on the images. The image is also projected on the model using the homography.}
     \label{fig:qual}
\end{figure}

\paragraph{\bf Qualitative Results:}
In Fig. \ref{fig:qual} we project the model on a few test images using the homography obtained with our best features (G+VerL+HorL+C). We also project the image on the model of the field. We observe great agreement between the image and the model.

\begin{figure}[t]
     \centering
\subfloat{
	\includegraphics[width=0.33\linewidth]{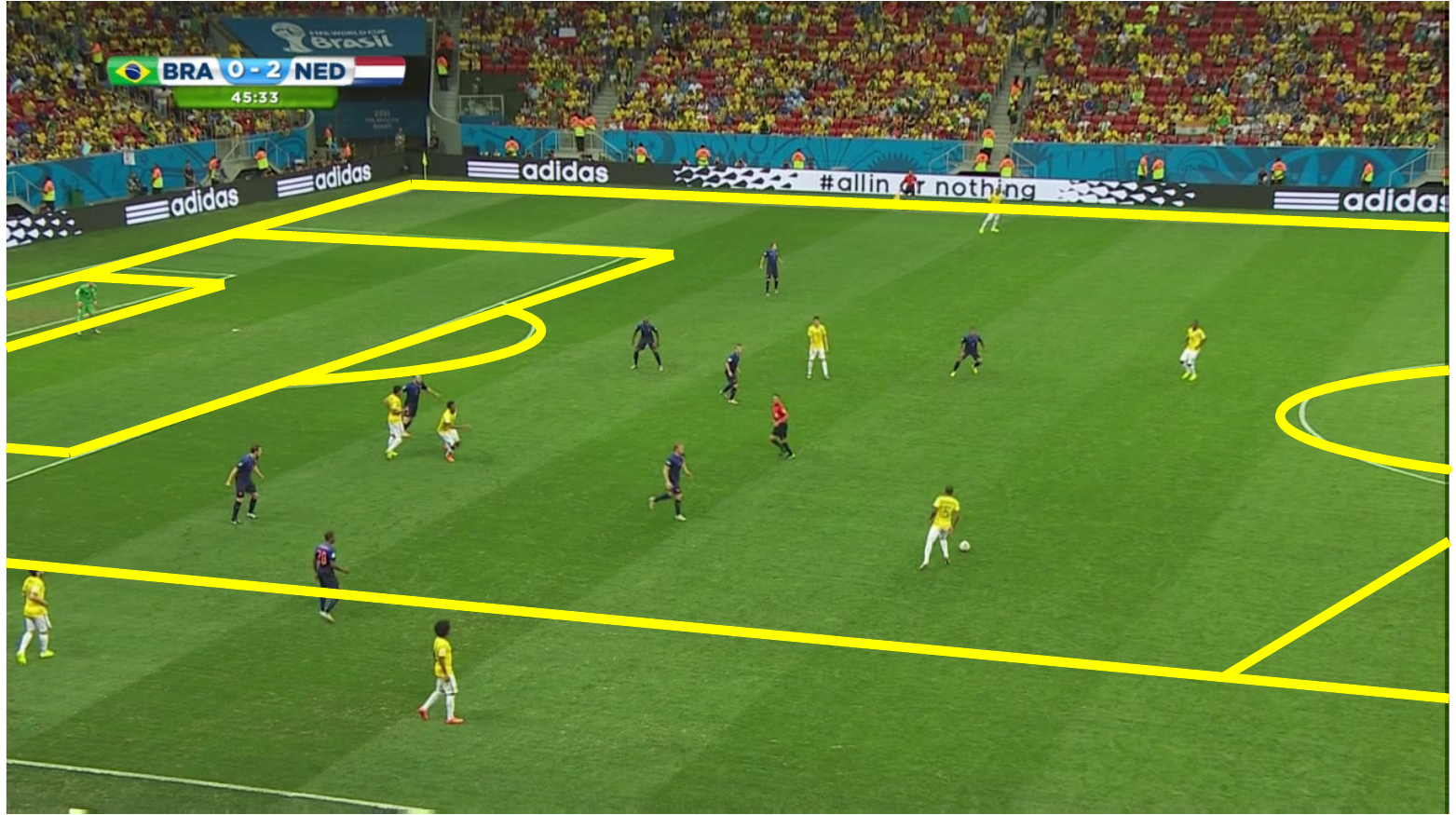}
	 }
\subfloat{
	\includegraphics[width=0.33\linewidth]{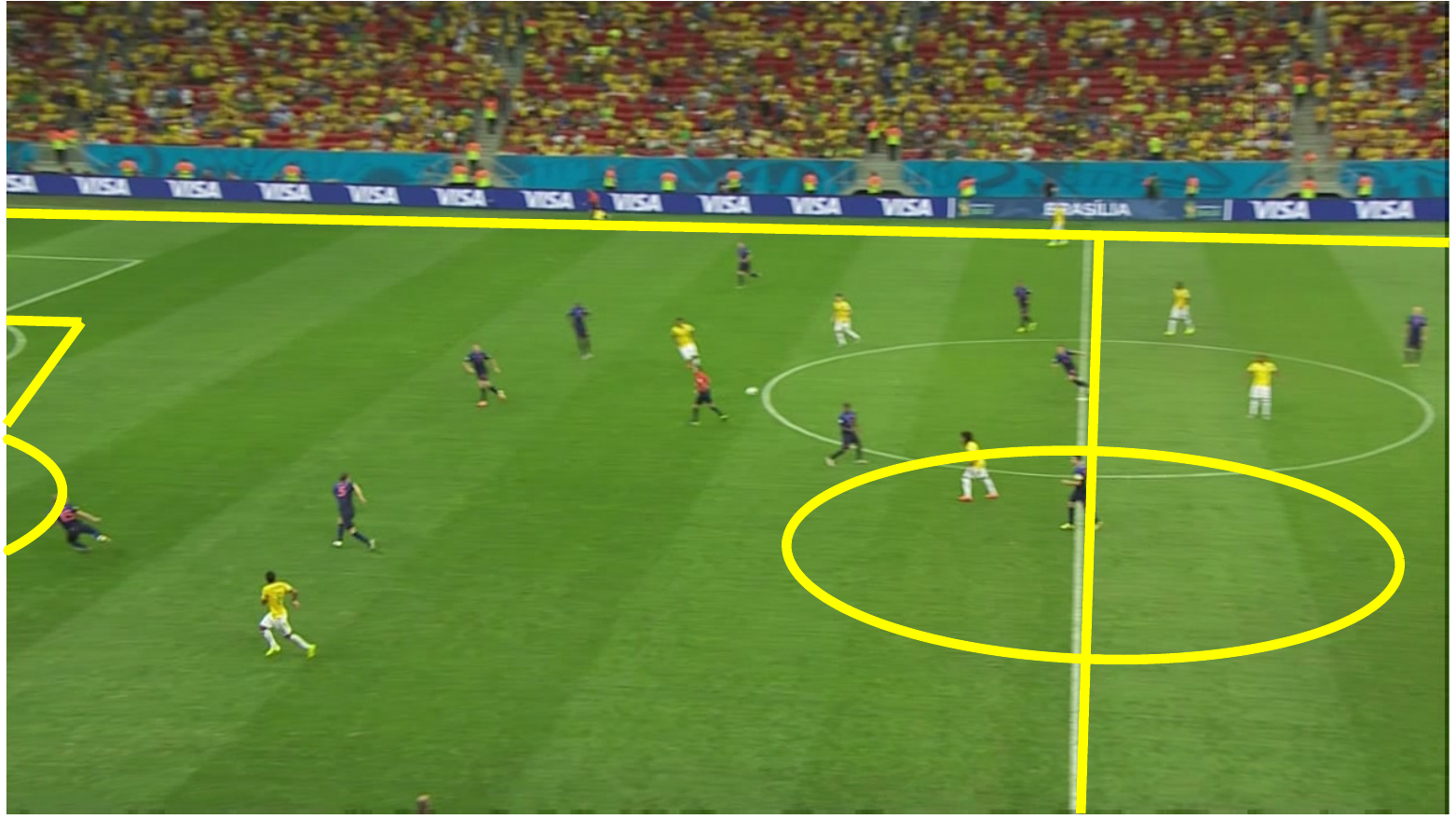}
  }
\subfloat{
       	\includegraphics[width=0.33\linewidth]{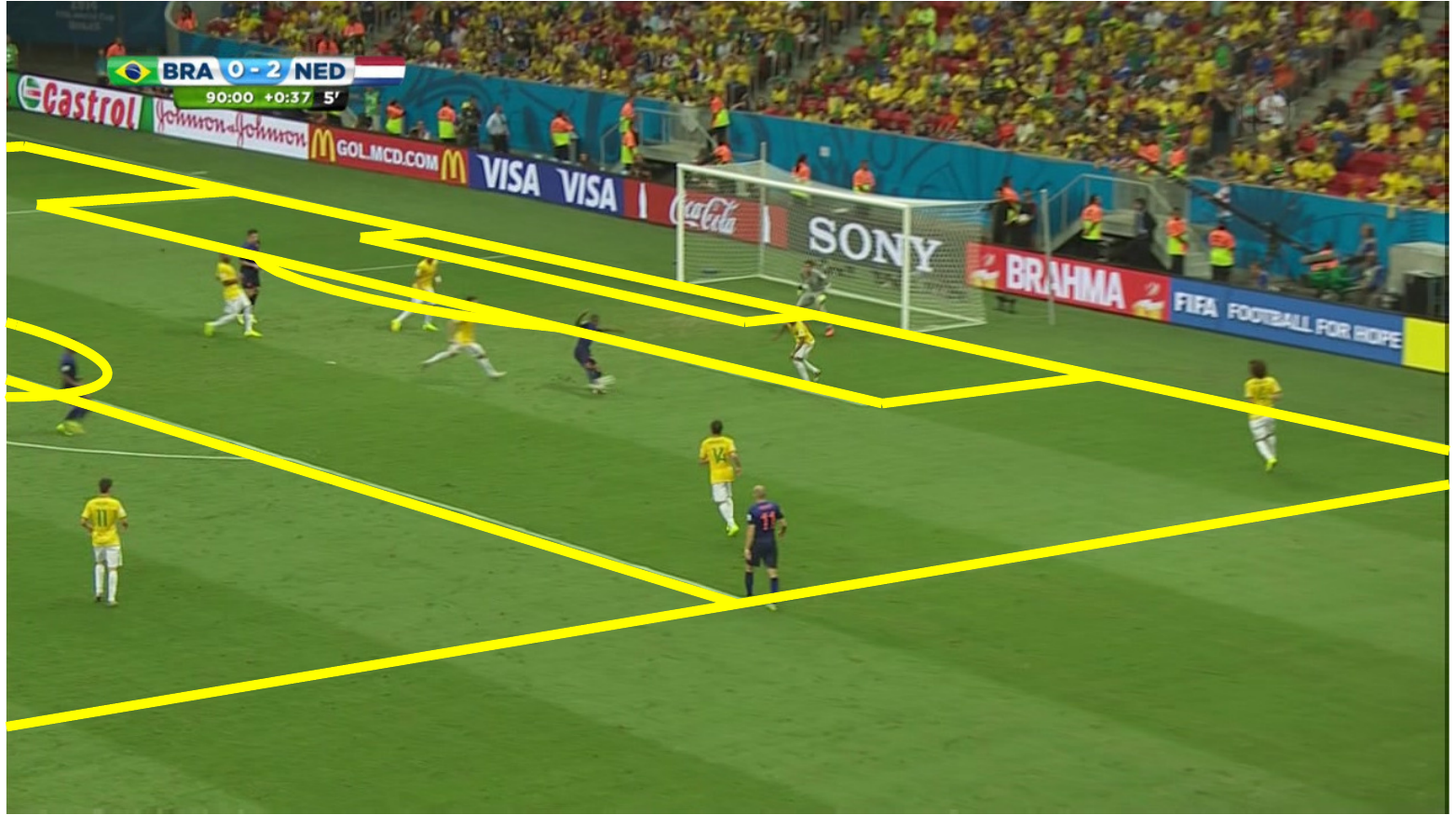}
  }
\\
\subfloat{
	\includegraphics[width=0.33\linewidth]{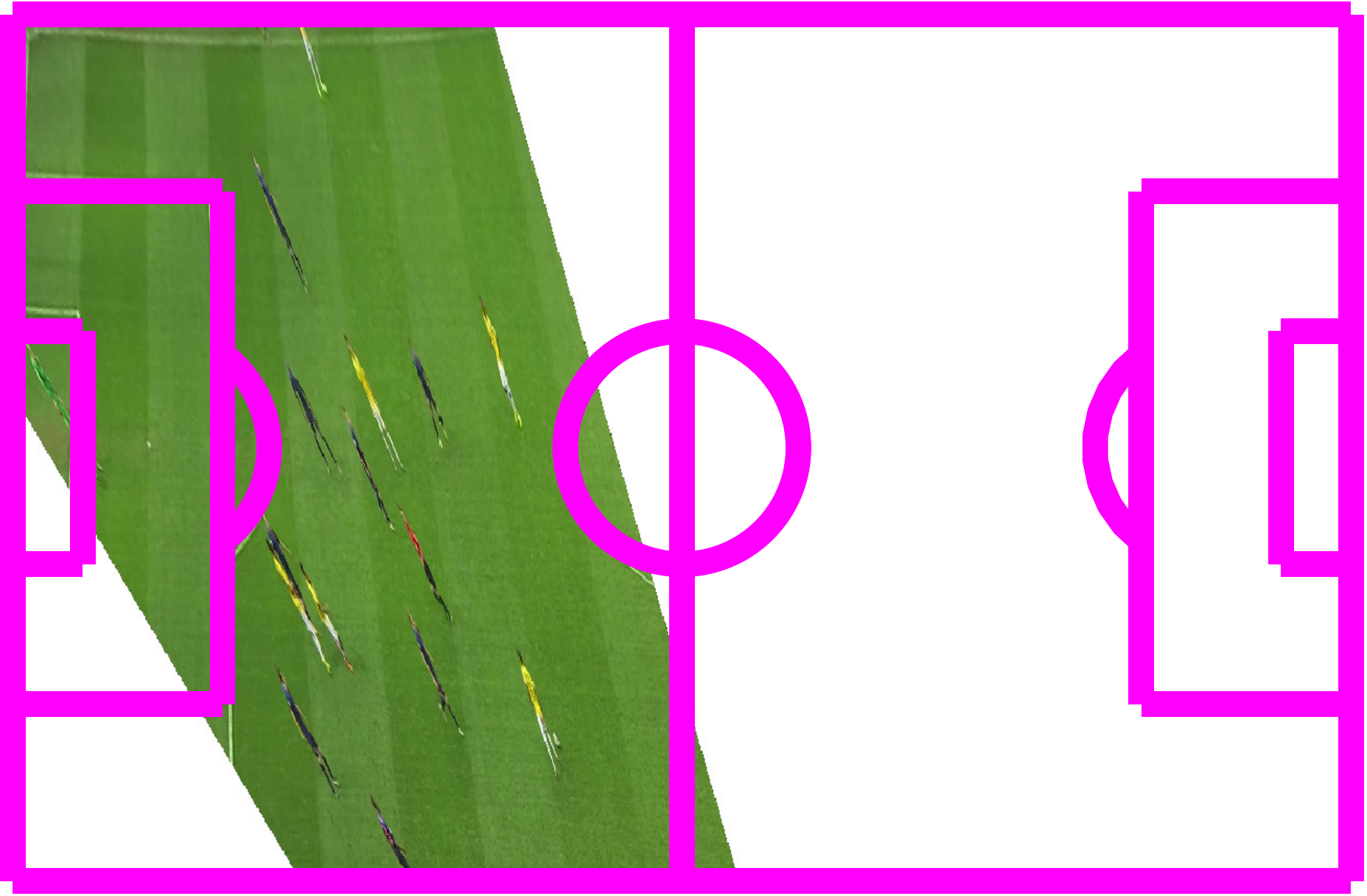}
	 }
\subfloat{
	\includegraphics[width=0.33\linewidth]{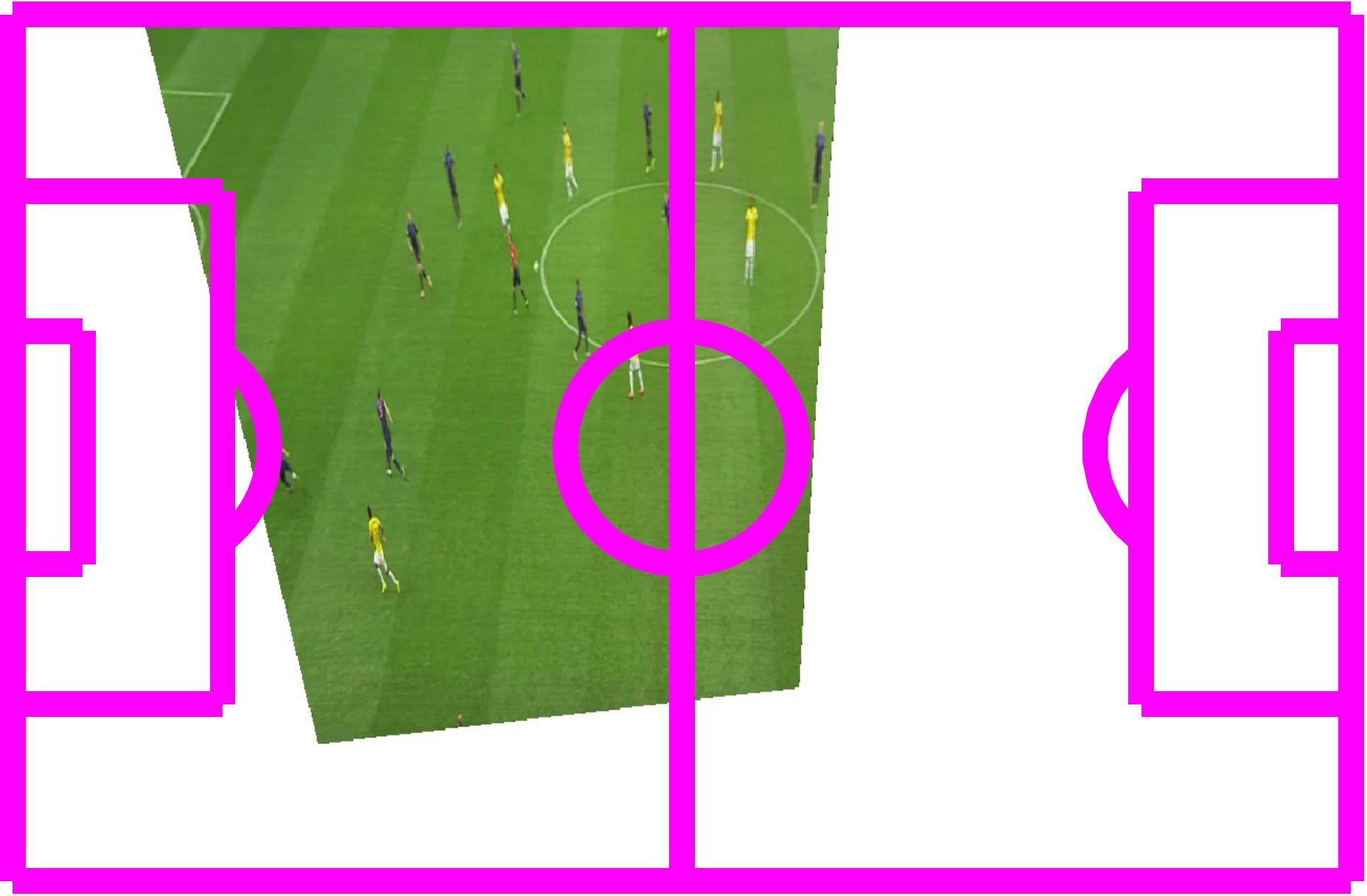}
  }
\subfloat{
       	\includegraphics[width=0.33\linewidth]{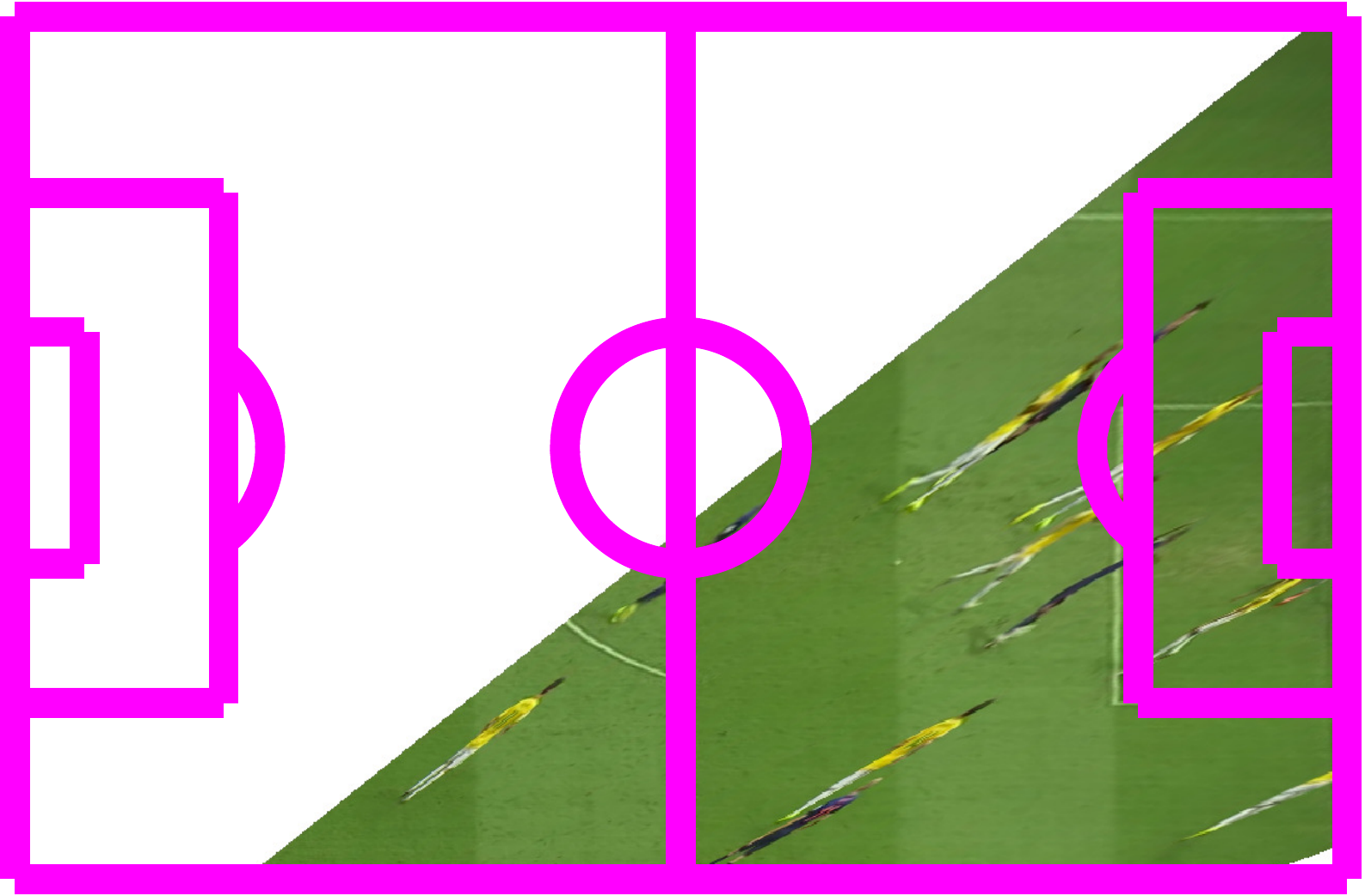}
  }

     \caption{Three failure examples where the homography is not correctly estimated}
     \label{fig:fail}
\end{figure}

\paragraph{\bf Failure Modes:} Fig. \ref{fig:fail} shows failure modes which are mainly due to errors due to failure of the circle potential.  

\paragraph{\bf Speed and Number of Iterations.}
For the best set of features (denoted with G+VerL+VerH+C in Table~\ref{table:ablation}), it takes on average 0.7 seconds (a median of 0.5) to perform inference and on average 2964 BBound iterations (with median of 1848 iterations). Times clocked on one core of AMD Opteron 6136. 

\section{Conclusion and Future Work}

In this paper, we presented a new framework for fast and automatic field localization as applied to the game of soccer. We framed this problem as a branch and bound inference task in a Markov Random Field. We evaluated our method on collection of broadcast images recorded from World Cup 2014. 
As was mentioned, we do not take into account temporal information in our energy function. For future work, we intend to construct temporal potential functions and evaluate our method on video sequences. We also plan to incorporate player detection and tracking in our framework. Finally, we aim to extend our method to other team sports such as hockey, basketball, rugby and American Football.


\bibliographystyle{splncs}
\bibliography{topArxiv}

\end{document}